\documentclass[runningheads]{llncs}
\usepackage{graphicx}

\usepackage{tikz}
\usepackage{comment}
\usepackage{amsmath,amssymb} 
\usepackage{color}
\usepackage{algorithm}
\usepackage{algpseudocode}
\usepackage{dirtytalk}
\usepackage{multirow}
\usepackage{soul}
\usepackage{caption}
\usepackage{subcaption}
\usepackage{colortbl}
\usepackage{enumitem}
\usepackage{url}
\usepackage{orcidlink}

\usepackage[accsupp]{axessibility}  

\newcommand{\cE}{\mathcal{E}}

\newcommand{\cS}{\mathcal{S}}

\newcommand{\bu}{\mathbf{u}}

\newcommand{\bP}{\mathbf{P}}

\newcommand{\bp}{\mathbf{p}}

\newcommand{\bV}{\mathbf{V}}
\newcommand{\bv}{\mathbf{v}}

\newcommand{\be}{\mathbf{e}}

\newcommand{\cV}{\mathcal{V}}

\newcommand*{\etc}{\emph{etc.}}
\newcommand*{\eg}{\emph{e.g.}}
\newcommand*{\ie}{\emph{i.e.}}
\newcommand*{\etal}{\emph{et al.}}

\begin{document}
\pagestyle{headings}
\mainmatter
\def\ECCVSubNumber{175}  

\title{Globally Optimal Event-Based Divergence Estimation for Ventral Landing} 

\titlerunning{Globally Optimal Event-Based Divergence Estimation for Ventral Landing}
%
\author{{Sofia McLeod}\inst{1}\index{McLeod, Sofia} \and
{Gabriele Meoni}\inst{2} \and
{Dario Izzo}\inst{2}\and
{Anne Mergy}\inst{2} \and
{Daqi Liu}\inst{1}\and
{Yasir Latif}\inst{1}\and
{Ian Reid}\inst{1}\and
{Tat-Jun Chin}\inst{1}
}
%
\authorrunning{S. McLeod et al.}
%
\institute{School of Computer Science, The University of Adelaide \email{@adelaide.edu.au} \and
Advanced Concepts Team, European Space Research and Technology Centre,
Keplerlaan 1, 2201 AZ Noordwijk, The Netherlands
\email{@esa.int}}
\maketitle

\begin{abstract}
Event sensing is a major component in bio-inspired flight guidance and control systems. We explore the usage of event cameras for predicting time-to-contact (TTC) with the surface during ventral landing. This is achieved by estimating divergence (inverse TTC), which is the rate of radial optic flow, from the event stream generated during landing. Our core contributions are a novel contrast maximisation formulation for event-based divergence estimation, and a branch-and-bound algorithm to exactly maximise contrast and find the optimal divergence value. GPU acceleration is conducted to speed up the global algorithm. Another contribution is a new dataset containing real event streams from ventral landing that was employed to test and benchmark our method. Owing to global optimisation, our algorithm is much more capable at recovering the true divergence, compared to other heuristic divergence estimators or event-based optic flow methods. With GPU acceleration, our method also achieves competitive runtimes.
\end{abstract}

\section{Introduction}\label{sec:intro}



Flying insects employ seemingly simple sensing and processing mechanisms for flight guidance and control. Notable examples include the usage of observed optic flow (OF) by honeybees for flight control and odometry~\cite{Srinivasan1996-na}, and observed OF asymmetries for guiding free flight behaviours in fruit flies~\cite{Tammero2002-hi}. Taking inspiration from insects to build more capable flying machines~\cite{Fry2010-qj,srinivasan2011,Srinivasan2010-by}, a branch of biomimetic engineering, is an active research topic.

The advent of event sensors~\cite{Brandli2014-rg,Lichtsteiner2008-mg,Posch2011-qh} has helped push the boundaries of biomimetic engineering. Akin to biological eyes, a pixel in an event sensor asynchronously triggers an event when the intensity change at that pixel exceeds a threshold. An (ideal) event sensor outputs a stream of events if it observes continuous changes, \eg, due to motion, else it generates nothing. The asynchronous pixels also enable higher dynamic range and lower power consumption. These qualities make event sensors attractive as flight guidance sensors for aircraft~\cite{Dinaux2021-lg,Falanga2020-go,Mueggler2014-tz,Pijnacker_Hordijk2018-xj,Sanket2020-wq,Vidal2018-hz} and spacecraft~\cite{Orchard2009-na,Sikorski2021-tf,Valette2010-ps,Vasco-uu}.

\begin{figure}[t]
     \centering
     \begin{subfigure}[b]{0.25\textwidth}
         \centering
         \includegraphics[width=\textwidth]{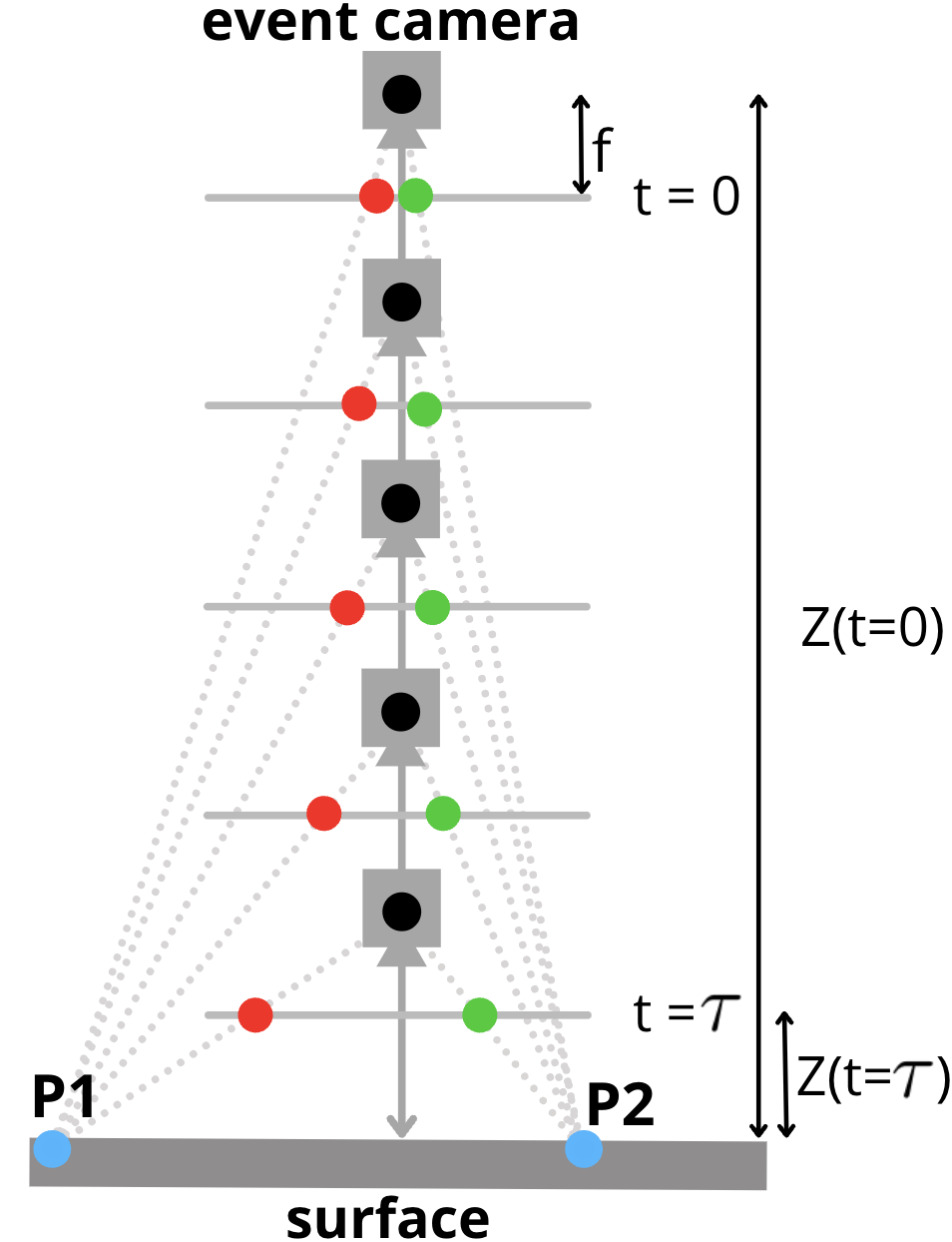}
         \caption{}
         \label{fig:vertical_landing}
     \end{subfigure}
     \hfill
     \begin{subfigure}[b]{0.3\textwidth}
         \centering
         \includegraphics[width=\textwidth]{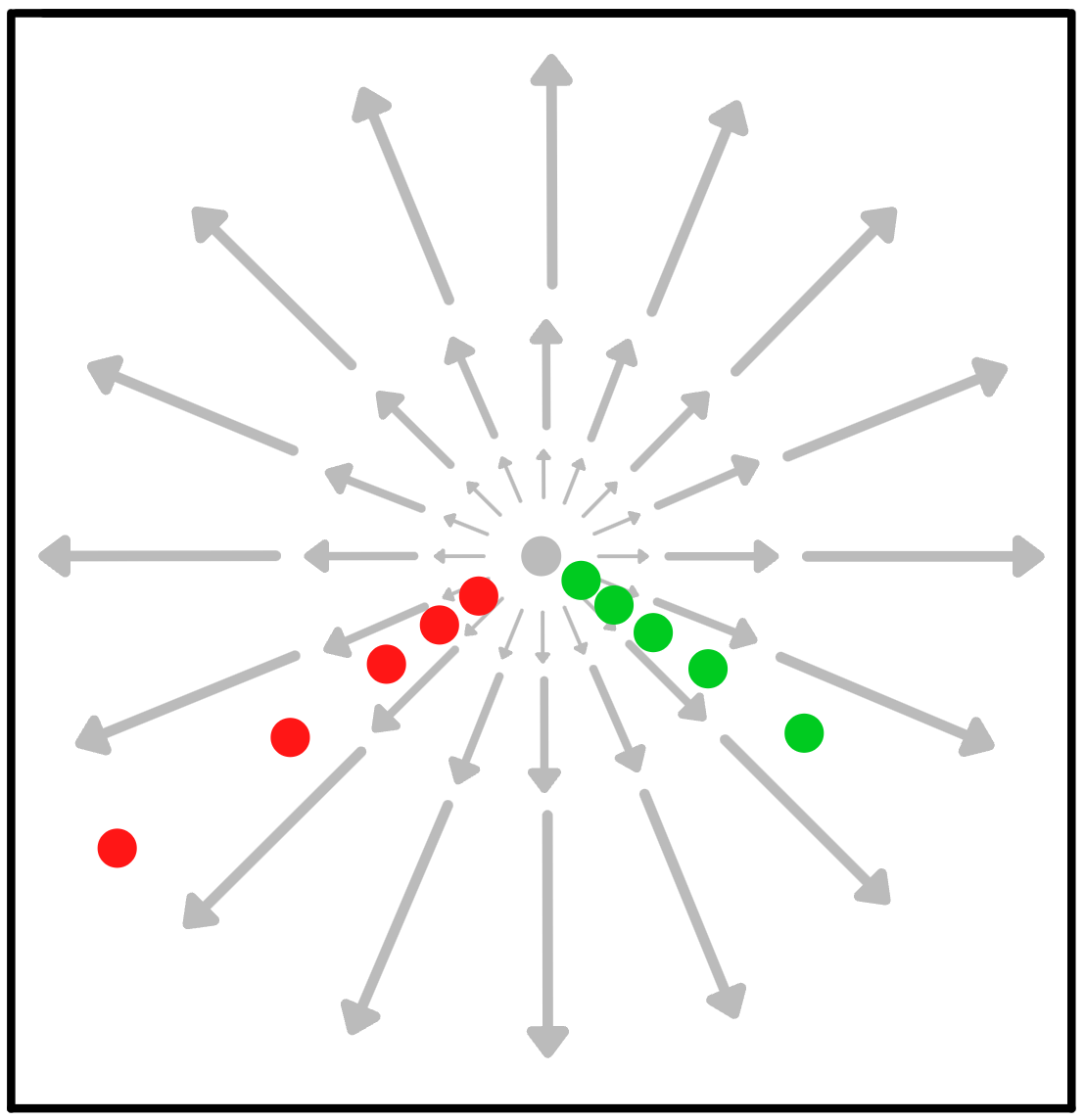}
         \caption{}
         \label{fig:expected_of}
     \end{subfigure}
     \hfill
     \begin{subfigure}[b]{0.4\textwidth}
         \centering
         \includegraphics[width=\textwidth]{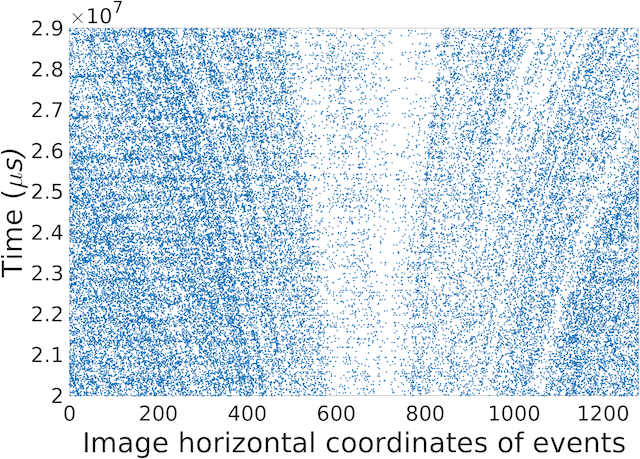}
         \caption{}
         \label{fig:event_trajectories}
     \end{subfigure}
        \caption{(a) An event camera with focal length $f$ undergoing ventral descent to a surface over time $0$ to $\tau$ seconds. (b) Two example event trajectories triggered by observation of scene points $P1$ and $P2$ on the surface. The event trajectories are asynchronous observations of the radial optic flows, which emanate from the focus of expansion (FOE). (c) Actual event data produced under ventral descent (only the image horizontal coordinates and event time stamps are shown here).}
        \label{fig:vertical_descent}
\end{figure}

In previous works, divergence is typically recovered from the OF vectors resulting from event-based OF estimation~\cite{Pijnacker_Hordijk2018-xj,Sikorski2021-tf,Valette2010-ps,Vasco-uu}. However, OF estimation is challenged by noise in the event data (see~Fig.~\ref{fig:event_image_no_velocity}), which leads to inaccurate divergence estimates (as we will show in Sec.~\ref{sec:results}). Fundamentally, OF estimation, which usually includes detecting and tracking features, is more complex than estimating divergence (a scalar quantity).

\subsubsection{Our contributions}

\begin{figure}[t]\centering
     \begin{subfigure}[b]{0.32\textwidth}
         \centering
         \includegraphics[width=\textwidth]{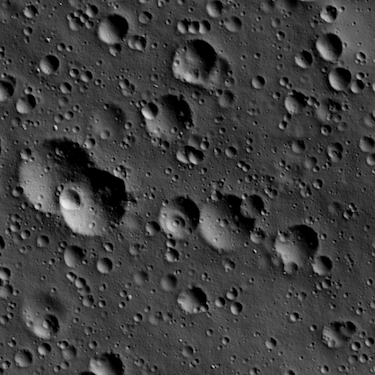}
         \caption{}
         \label{fig:rgb_img_vertical_descent}
     \end{subfigure}
     \hfill     
     \begin{subfigure}[b]{0.32\textwidth}
         \centering
         \includegraphics[width=\textwidth]{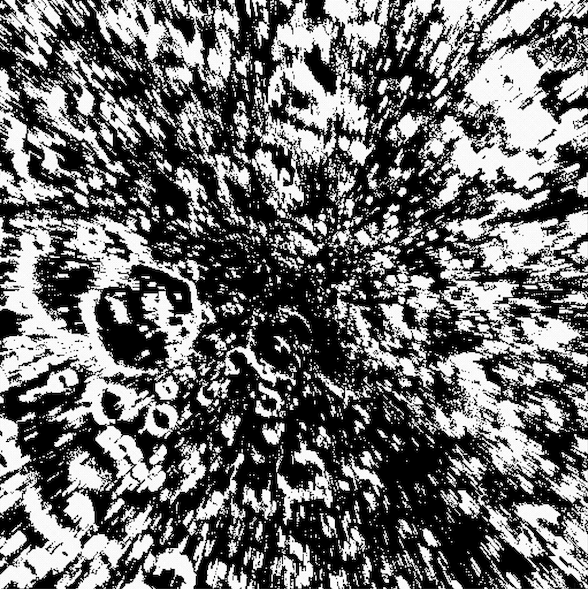}
         \caption{}
         \label{fig:event_image_no_velocity}
     \end{subfigure}
     \hfill
     \begin{subfigure}[b]{0.32\textwidth}
         \centering
         \includegraphics[width=\textwidth]{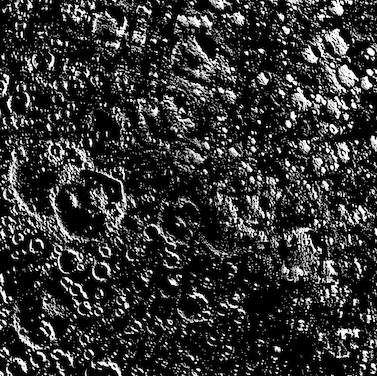}
         \caption{}
         \label{fig:event_image_true_velocity}
     \end{subfigure}
        \caption{(a) Intensity image of landing surface. (b) Event image (without motion compensation) for a batch of events generated while observing the surface during vertical landing (image contrast is 0.15). (c) Motion compensated event image produced by our divergence estimation method (image contrast is 0.44).}
        \label{fig:vertical_descent_examples}
\end{figure}

We present a \emph{contrast maximisation}~\cite{Gallego2018-tu,Stoffregen2019-jc} formulation for event-based divergence estimation. We examined the geometry of ventral landing and derived event-based radial flow, then built a mathematically justified optimisation problem that aims to maximise the contrast of the flow-compensated image; see Fig.~\ref{fig:vertical_descent_examples}. To solve the optimisation, we developed an exact algorithm based on branch-and-bound (BnB)~\cite{Horst1996-xx}, which is accelerated using GPU.

Since our aim is divergence estimation, our mathematical formulation differs significantly from the exact methods of~\cite{Liu2020-eo,Peng2020-pw}, who focused on different problems. More crucially, our work establishes the practicality of exact contrast maximisation through GPU acceleration, albeit on a lower dimensional problem.

To test our method, we collected real event data from ventral landing scenarios (Sec.~\ref{sec:datacollection}), where ground truth values were produced through controlled recording using a robot arm and depth sensing\footnote{Dataset: \url{https://github.com/s-mcleod/ventral-landing-event-dataset}}. Results in Sec.~\ref{sec:results} show that our method is much more accurate than previous divergence estimators~\cite{Pijnacker_Hordijk2018-xj,Sikorski2021-tf} that conduct heuristic OF recovery (plane fitting, centroid tracking). Our method also exhibited superior accuracy over state-of-the-art event-based OF methods~\cite{Gehrig2021-ej,Stoffregen2018-nm,Stoffregen2019-jc} that we employed for divergence estimation. The usage of GPU acceleration enabled our global method to achieve competitive runtimes.

\section{Related work}\label{sec:related}


\subsubsection{Event-based optic flow}

Event-based OF is a rapidly growing area in computer vision.  Some event-based OF approaches~\cite{Bardow2016-di,Pan2020-ju} are based on the traditional brightness constancy assumption.  Other methods~\cite{Benosman2012-ud,Gallego2018-tu} expand on the Lucas and Kanade algorithm \cite{Lucas1981-yk}, which assumes constant, localised flow.  Alternative approaches include pure event-based time oriented OF estimations~\cite{Benosman2014-da}, computing distance transforms as input to intensity-based OF algorithms~\cite{Almatrafi2020-ot}, and adaptive block matching where OF is computed a points where brightness changes~\cite{Liu2018-bs}. An extension of event-based OF is event-based motion segmentation~\cite{Stoffregen2019-me,Stoffregen2019-jc,Zhou2020-cc}, where motion models are fitted to clusters of events that have the same OF.

Learning methods for event-based OF train deep neural networks (DNNs) to output event image OF estimations from input representations of event point clouds. Some of these methods require intensity frames in addition to the event data~\cite{Ding2021-so,Zhu2018-jh,Zhu2019-aa}, whereas others require event data only~\cite{Gehrig2019-vc,Gehrig2021-ej,Ye2018-my}. Closely relevant are image and video reconstruction from event methods~\cite{Paredes-Valles2021-xq,Rebecq2019-sz,Rebecq2019-lr,Scheerlinck2020-tg}, which employ frame-based and event-based OF techniques. Note that not all event sensors provide intensity frames and event streams.

\subsubsection{Event-based divergence estimation}


Using event cameras for flight guidance (\eg, attitude estimation, aerial manoeuvres) is an active research area~\cite{Chin2019-pq,Clady2014-vm,Gomez_Eguiluz2020-tc,Orchard2009-na}. The focus of our work is event-based divergence estimation, which is useful for guiding the ventral descent of aircraft~\cite{Pijnacker_Hordijk2018-xj} and spacecraft~\cite{Sikorski2021-tf}.


Pijnacker-Hordijk \textit{et al.}~\cite{Pijnacker_Hordijk2018-xj} employed and extended the plane fitting method of~\cite{Benosman2014-da} for event-based divergence estimation. The algorithm was also embedded in a closed-loop vertical landing system of an MAV. In the context of planetary landing, Sikorski~\textit{et al.}~\cite{Sikorski2021-tf} presented a local feature matching and tracking method, which was also embedded in a closed-loop guidance system. However, their system was evaluated only on simulated event data~\cite{Sikorski2021-tf}.




Since our focus is on the estimation of divergence, we will distinguish against the state-of-the-art methods~\cite{Pijnacker_Hordijk2018-xj,Sikorski2021-tf} only in that aspect, \ie, independently of a close-loop landing system. As we will show in Sec.~\ref{sec:results}, our globally optimal contrast maximisation approach provides much more accurate estimates than~\cite{Pijnacker_Hordijk2018-xj,Sikorski2021-tf}.

\subsubsection{Spiking neural networks}

Further pushing the boundaries of biomimetic engineering is the usage of spiking neural networks (SNN)~\cite{Orchard2013-wi} for event-based OF estimation~\cite{Haessig2018-gx,Hagenaars2021-lz,Lee2020-yb,Paredes-Valles2020-ej}. Due to the limited availability of SNN hardware~\cite{intel_lohi}, we consider only off-the-shelf computing devices (CPUs and GPUs) in our work, but we note that SNNs are attractive future research.


\section{Geometry of ventral landing}

We first examine the geometry of ventral descent and derive continuous radial OF and divergence, before developing the novel event-based radial flow.



\subsection{Continuous-time optic flow}


Let $\textbf{P}(t) = \left[ \begin{matrix} X(t), & Y(t), & Z(t) \end{matrix} \right]^T$ be a 3D scene point that is moving with velocity $\bV(t) = \left[ \begin{matrix} V_X(t), & V_Y(t), & V_Z(t) \end{matrix} \right]^T$. Assuming a calibrated camera, projecting $\bP(t)$ onto the image plane yields a moving image point
\begin{equation}\label{point_1}
    \textbf{p}(t) = \begin{bmatrix} x(t), & y(t) \end{bmatrix}^T := \left[ \begin{matrix} f X(t)/Z(t), & f Y(t)/Z(t) \end{matrix} \right]^T,
\end{equation}
where $f$ is the focal length of the event camera. Differentiating $\bp(t)$ yields the \emph{continuous-time OF} of the point, \ie,
\begin{equation}\label{of_1}
    \textbf{v}(t) = \begin{bmatrix} \dfrac{fV_X(t)-V_Z(t)x(t)}{Z(t)}, & \dfrac{fV_Y(t)-V_Z(t)y(t)}{Z(t)} \end{bmatrix}.
\end{equation}
Note that the derivations above are standard; see, \eg,~\cite[Chapter 10]{Forsyth2011-dj}.



\subsection{Continuous-time radial flow and divergence}\label{sec:assumptions}

Following~\cite{Pijnacker_Hordijk2018-xj,Sikorski2021-tf}, we consider the ventral landing scenario depicted in Fig.~\ref{fig:vertical_landing}. The setting carries the following assumptions:
\begin{itemize}[leftmargin=1em,itemsep=0pt,parsep=0pt,topsep=2pt]
    \item All scene points lie on a fronto-parallel flat surface; and
    \item The camera has no horizontal motion, \ie, $V_X(t) = V_Y(t) = 0$, which implies that $\textbf{P}(t) = \left[ \begin{matrix} X, & Y, & Z(t) \end{matrix} \right]^T$, \ie, the first two elements of $\bP(t)$ are constants.
\end{itemize}
Sec.~\ref{sec:results} will test our ideas using real data which unavoidably deviated from the assumptions. Under ventral landing, \eqref{of_1} is simplified to
\begin{equation}\label{of_2}
    \textbf{v}(t) = -\left[V_Z(t)/Z(t)\right]\bp(t).
\end{equation}
Since the camera is approaching the surface, $V_Z(t) \le 0$, hence $\bv(t)$ points away from $\mathbf{0}$ (the FOE) towards the direction of $\bp(t)$; see Fig.~\ref{fig:expected_of}.


The \emph{continuous-time divergence} is the quantity
\begin{align}
D(t) = V_Z(t)/Z(t),
\end{align}
which is the rate of expansion of the radial OF. The divergence is independent of the scene point $\bP(t)$, and the inverse of $D(t)$, \ie, distance to surface over rate of approach to the surface, is the \emph{instantaneous TTC}.



\subsection{Event-based radial flow}\label{sec:ebrf}


Let $\cS$ be an event stream generated by an event camera during ventral descent. Following previous works, \eg~\cite{Gallego2018-tu,Pan2020-ju,Stoffregen2019-me}, we separate $\cS$ into small finite-time sequential batches. Let $\cE = \{\be_i\}^{N}_{i=1} \subset \cS$ be a batch with $N$ events, where each $\be_i = (x_i,y_i,t_i,p_i)$ contains image coordinates $(x_i,y_i)$, time stamp $t_i$ and polarity $p_i$ of the $i$-th event. To simplify exposition, we offset the time by conducting $t_i = t_i - \min_i t_i$ such that the period of $\cE$ has the form $[0,\tau]$.

Since $\tau$ is small relative to the speed of descent, we assume that $V_Z(t)$ is constant in $[0,\tau]$. We can thus rewrite~\eqref{of_2} corresponding to $\bP(t)$ as
\begin{equation}\label{of_4}
    \textbf{v}(t) = \dfrac{-\nu}{Z_0+\nu t}\begin{bmatrix} \dfrac{fX}{Z_0+\nu t}, & \dfrac{fY}{Z_0+\nu t}  \end{bmatrix}^T, \;\;\;\;  \forall t \in [0,\tau],
\end{equation}
where $\nu$ is the constant vertical velocity in $[0,\tau]$, and $Z_0$ is the surface depth at time $0$ (recall also that the horizontal coordinates of $\bP(t)$ are constant under ventral descent). Accordingly, divergence is redefined as
\begin{align}\label{eq:batchdiv}
D(t) = \nu/(Z_0 + \nu t).
\end{align}

Assume that $\be_i$ was triggered by scene point $\bP(t)$ at time $t_i \in [0,\tau]$, \ie, $\be_i$ is a point observation of the continuous flow $\bv(t)$. This means we can recover
\begin{equation}\label{rw_1}
X = x_i(Z_0+\nu t_i)/f, \;\;\;\; Y = y_i(Z_0+\nu t_i)/f,
\end{equation}
following from~\eqref{point_1}. Substituting~\eqref{rw_1} into~\eqref{of_4} yields
\begin{equation}\label{of_5}
    \textbf{v}(t) = \dfrac{-\nu}{Z_0+\nu t}\begin{bmatrix} \dfrac{x_i(Z_0+\nu t_i)}{Z_0+\nu t}, & \dfrac{y_i(Z_0+\nu t_i)}{Z_0+\nu t}  \end{bmatrix}^T,
\end{equation}
The future event triggered by $\bP(t)$ at time $t^\prime$, where $t_i \le t^\prime \le \tau$, is
\begin{equation}\label{warp}
    \bp^\prime = \begin{bmatrix} x_i \\ y_i \end{bmatrix} + \int_{t_i}^{t^\prime} \textbf{v}(t) dt = \begin{bmatrix} x_i \\ y_i \end{bmatrix} + \Gamma_i(t^\prime),
\end{equation}
and $\Gamma_i(t^\prime)$ is called the \emph{event-based radial flow}
\begin{equation}\label{warp_expanded}
    \Gamma_i(t^\prime) := \begin{bmatrix} \dfrac{x_i(Z_0+\nu t_i)}{Z_0+\nu t^\prime} - x_i, & \dfrac{y_i(Z_0+\nu t_i)}{Z_0+\nu t^\prime} - y_i \end{bmatrix}^T.
\end{equation}
While~\eqref{warp} seems impractical since it requires knowing the depth $Z_0$, we will circumvent \emph{a priori} knowledge of the depth next.

\section{Exact algorithm for divergence estimation}\label{sec:algo}

Following the batching procedure in Sec.~\ref{sec:ebrf}, our problem is reduced to estimating divergence~\eqref{eq:batchdiv} for each event batch $\cE$ with assumed time duration $[0,\tau]$. Here, we describe our novel algorithm to perform the estimation exactly.

\subsection{Optimisation domain and retrieval of divergence}\label{sec:domain}

There are two unknowns in~\eqref{eq:batchdiv}: $Z_0$ and $\nu$. Since the event camera approaches the surface during descent, we must have $\nu \le 0$. We also invoke cheirality, \ie, the surface is always in front of the camera, thus $Z_0 + \nu \tau \ge 0$. These imply
\begin{align}
    -Z_0/\tau \le \nu \le 0.
\end{align}
Note also that $D(t)$ is a ratio, thus the precise values of $Z_0$ and $\nu$ are not essential. We can thus fix, say, $Z_0 = 1$, and scale $\nu$ accordingly to achieve the same divergence. The aim is now reduced to estimating $\nu$ from $\cE$.

Given the estimated $\nu^\ast$, the divergence can be retrieved as $D(t) = \nu^\ast/(1 + \nu^\ast t)$. In our experiments, we take the point estimate $D^\ast = \nu^\ast/(1 + \nu^\ast \tau)$.

\begin{figure}[t]\centering
      \begin{subfigure}[b]{0.55
      \textwidth}
         \centering
         \includegraphics[width=\textwidth]{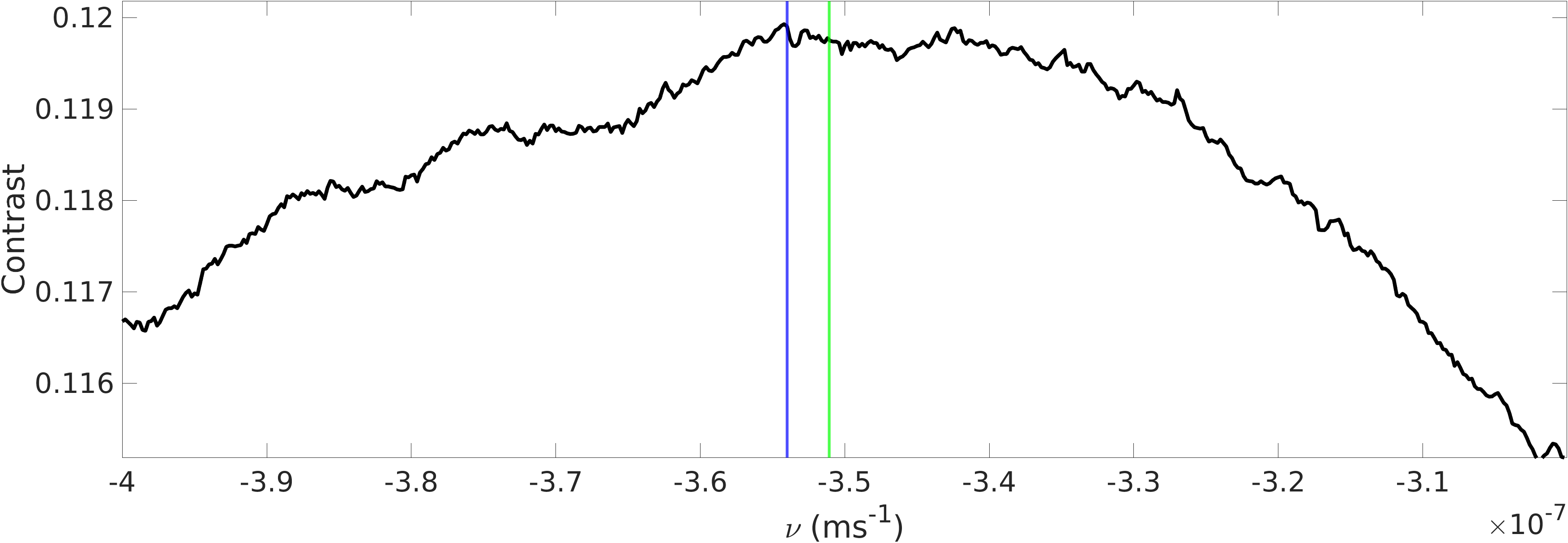}
         \caption{}
         \label{fig:plot_div_vs_contrast}
     \end{subfigure}
     \hfill
     \begin{subfigure}[b]{0.21\textwidth}
         \centering
         \includegraphics[width=\textwidth]{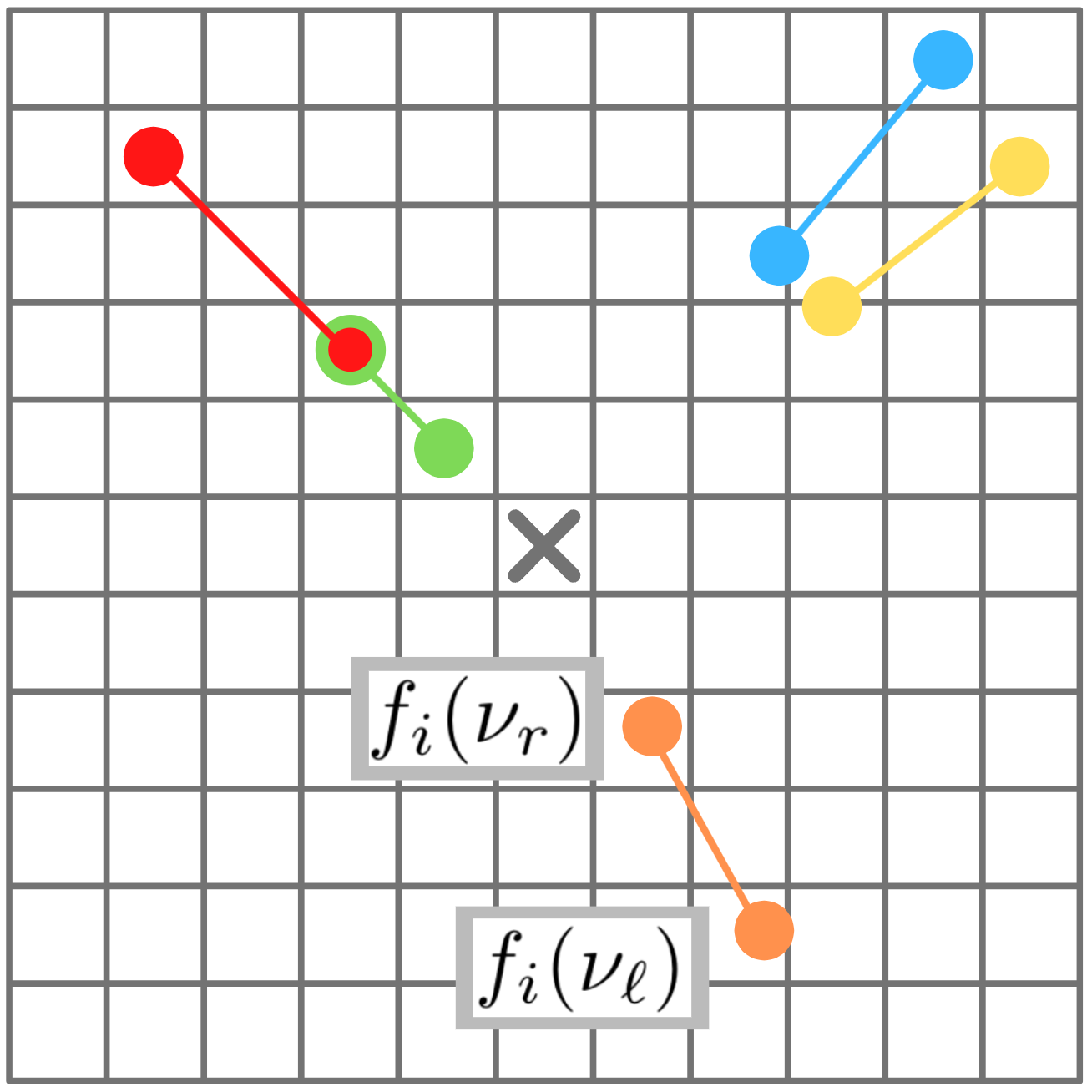}
         \caption{}
         \label{fig:ub_events}
     \end{subfigure}
     \hfill     
     \begin{subfigure}[b]{0.21\textwidth}
         \centering
         \includegraphics[width=\textwidth]{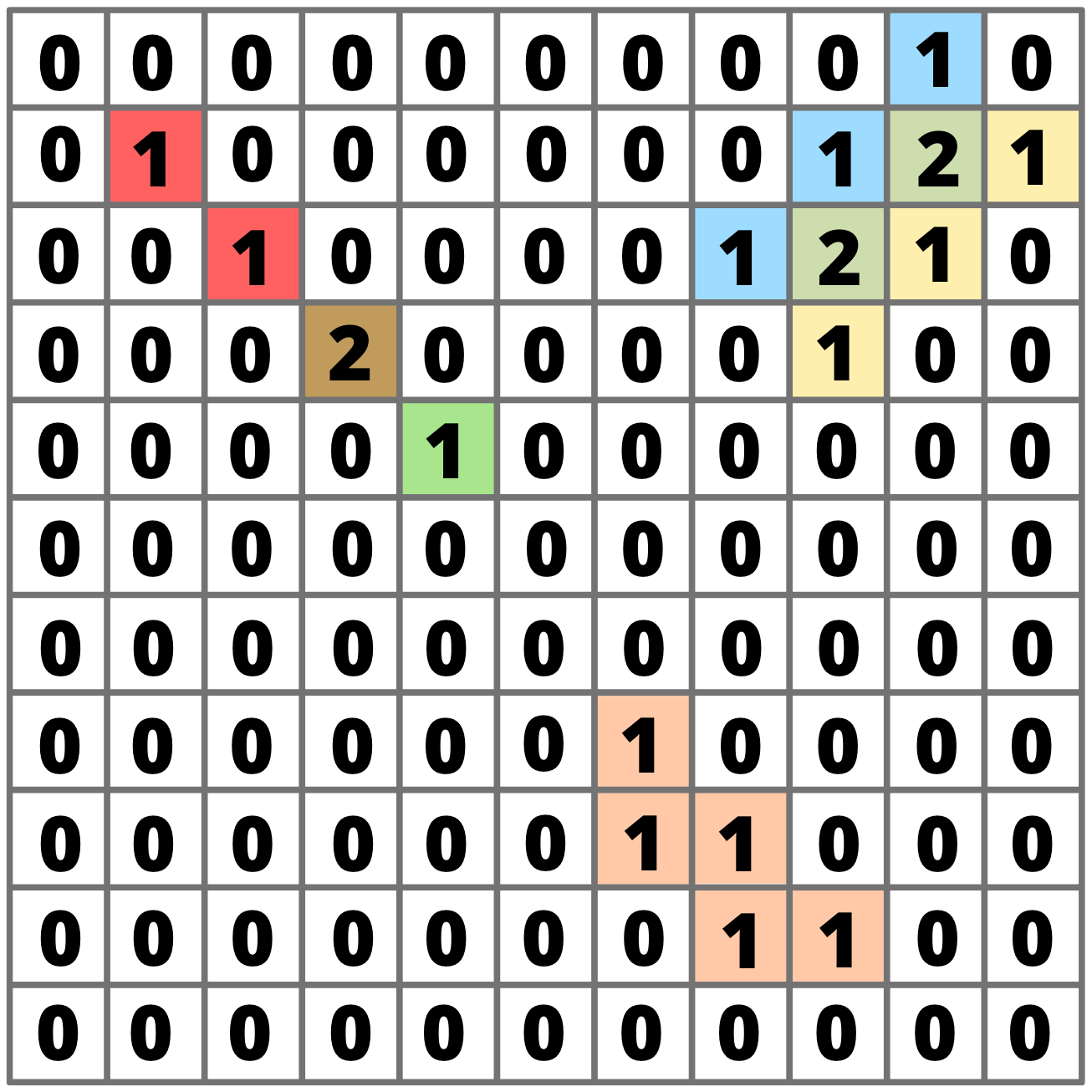}
         \caption{}
         \label{fig:ub_pixels}
     \end{subfigure}
        \caption{(a) Plot of the contrast function~\eqref{contrast} for an event batch. Blue and green lines represent optimal and ground truth velocities respectively. Fig.~\ref{fig:event_image_true_velocity} was produced by the optimal velocity. (b) Event lines $f_i(\nu_\ell) \leftrightarrow f_i(\nu_r)$ corresponding to five events (differentiated by colours) under velocity domain $\cV = [\nu_\ell,\nu_r]$. (c) Upper bound image $\bar{H}$ whose computation is GPU accelerated.}


        \label{fig:upper_bound}
\end{figure}

\subsection{Contrast maximisation}

To estimate $\nu$ from $\cE = \{\be_i\}^{N}_{i=1}$, we define the \emph{event-based radial warp}
\begin{equation}\label{eq:warping_function}
    f_i(\nu) = \begin{bmatrix} x_i \\ y_i \end{bmatrix} + \begin{bmatrix} \dfrac{x_i(Z_0+\nu t_i)}{Z_0+\nu \tau} - x_i, & \dfrac{y_i(Z_0+\nu t_i)}{Z_0+\nu \tau} - y_i \end{bmatrix}^T,
\end{equation}
which is a specialisation of~\eqref{warp} by grounding $Z_0 = 1$ and warping $\be_i$ to the end $\tau$ of the duration of $\cE$. The motion compensated image
\begin{equation}\label{image_plane}
    H(\bu;\nu)= \sum_{i=1}^{N} \mathbb{I} (f_i(\nu) \textrm{ lies in pixel } \bu),
\end{equation}
accumulates events that are warped to the same pixel (note that we employ the discrete version of~\cite{Liu2020-eo}). Intuitively, noisy events triggered by the same scene point will be warped by $f_i(\nu)$ to close-by pixels. This motivates contrast~\cite{Gallego2018-tu}
\begin{equation}\label{contrast}
    C(\nu) = \dfrac{1}{M}\sum_{\bu} (H(\bu;\nu) - \mu (\nu))^2,
\end{equation}
of the motion compensated image as an objective function for estimating $\nu$, where $M$ is the number of pixels in the image, and $\mu(\nu)$ is the mean of $H(\bu; \nu)$. Fig.~\ref{fig:plot_div_vs_contrast} illustrates the suitability of~\eqref{contrast} for our aim. In particular, the true $\nu^\ast$ is very close to the maximiser of the contrast. Note that, following~\cite{Gallego2018-tu,Liu2020-eo} we do not utilise the event polarities, but they can be accommodated if desired.

\subsection{Exact algorithm}

As shown in Fig.~\ref{fig:plot_div_vs_contrast}, contrast~\eqref{contrast} is non-convex and non-smooth. Liu \etal~\cite{Liu2020-eo} demonstrated that maximising the contrast using approximate methods (\eg, via smoothing and gradient descent) could lead to bad local optima. Inspired by~\cite{Liu2020-eo,Peng2020-pw}, we developed a BnB method (Algorithm~\ref{alg:cap}) to exactly maximise~\eqref{contrast} over the domain $\cV = [-1/\tau, 0]$ of the vertical velocity $\nu$.

Starting with $\cV$, Algorithm~\ref{alg:cap} iteratively splits and prunes $\cV$ until the maximum contrast is found. For each subdomain, the upper bound $\bar{C}$ of the contrast over the subdomain (more details below) is computed and compared against the contrast of the current best estimate $\hat{\nu}$ to determine whether the subdomain can be discarded. The process continues until the difference between current best upper bound and the quality of $\hat{\nu}$ is less than a predefined threshold $\gamma$. Then, $\hat{\nu}$ is the globally optimal solution, up to $\gamma$. Given the velocity estimate from Algorithm~\ref{alg:cap}, Sec.~\ref{sec:domain} shows how to retrieve the divergence.

\begin{algorithm}[t]
\caption{Exact contrast maximisation for ventral velocity estimation.}\label{alg:cap}
\begin{algorithmic}[1]
\Require Event batch $\cE = \{\be_i\}_i^N$, batch duration $[0,\tau]$, convergence threshold $\gamma$.
\State Initialise $\cV \gets [-1/\tau,0]$, $\hat{\nu} \gets $centre of $\cV$.
\State $\textrm{Insert } \cV \textrm{ into priority queue}~Q \textrm{ with priority } \Bar{C}(\cV)$.
\While{$Q \textrm{ is not empty}$}
\State $\cV \gets \textrm{dequeue first item from } Q$.
\State \textbf{if} $\Bar{C}(\cV) - C(\hat{\nu}) \leq \gamma$ \textbf{then} \textbf{break}.
\State $\nu_c \gets $ centre of $\cV$.
\State \textbf{if} $C(\nu_c) \geq C(\hat{\nu})$ \textbf{then} $\hat{\nu} \gets \nu_c$.
\State Split $\cV$ equally into 2 subdomains $\cV_1$ and $\cV_2$.
\For{$s=1,2$}
    \State \textbf{if} $\Bar{C}(\cV_s) \geq C(\hat{\nu})$ \textbf{then} insert $\cV_s$ into $Q$ with priority $\Bar{C}(\cV_s)$.
\EndFor
\EndWhile
\State \textbf{return} $\hat{\nu}$.
\end{algorithmic}
\end{algorithm}

\subsubsection{Upper bound function}\label{sec:ub}


Following~\cite[Sec.~3]{Liu2020-eo}, we use the general form
\begin{align}\label{eq:daqiform}
    \bar{C}(\cV) = \frac{1}{M}\bar{S}(\cV) - \underline{\mu}(\cV)^2,
\end{align}
as the upper bound for contrast, where the terms respectively satisfy
\begin{align}\label{eq:sos}
    \bar{S}(\cV) \ge \max_{\nu \in \cV} \sum_{\bu} H(\bu; \nu)^2 \;\;\;\; \textrm{and} \;\;\;\; \underline{\mu}(\cV) \le \min_{\nu \in \cV} \mu(\nu),
\end{align}
as well as equality in~\eqref{eq:sos} when $\cV$ is singleton. See~\cite[Sec.~3]{Liu2020-eo} on the validity of~\eqref{eq:daqiform}. For completeness, we also re-derive~\eqref{eq:daqiform} in the supplementary material. 

It remains to specify $\bar{S}$ and $\underline{\mu}$ for our ventral landing case. To this end, we define the upper bound image
\begin{align}\label{eq:imgup}
    \bar{H}(\bu; \cV) = \sum_{i=1}^N \mathbb{I}(f_i(\nu_\ell) \leftrightarrow f_i(\nu_r) \textrm{ intersects with pixel } \bu),
\end{align}
where $\nu_\ell = \min(\cV)$ and $\nu_r = \max(\cV)$, and $f_i(\nu_\ell) \leftrightarrow f_i(\nu_r)$ is the line on the image that connects points $f_i(\nu_\ell)$ and $f_i(\nu_r)$; see Figs.~\ref{fig:ub_events} and~\ref{fig:ub_pixels}. We can establish
\begin{align}\label{eq:imguppbound}
    \bar{H}(\bu; \cV) \ge H(\bu; \nu), \;\;\;\; \forall \nu \in \cV,
\end{align}
with equality achieved when $\cV$ reaches a single point $\nu$ in the limit; see supplementary material for the proof. This motivates to construct
\begin{align}
    \bar{S}(\cV) = \sum_{\bu} \bar{H}(\bu; \cV)^2,
\end{align}
which satisfies the 1st condition in~\eqref{eq:sos}. For the 2nd term in~\eqref{eq:daqiform}, we use
\begin{align}\label{eq:mulower}
    \underline{\mu}(\cV) = \dfrac{1}{M}\sum_{i=1}^N\mathbb{I} (f_i(\nu_\ell) \leftrightarrow f_i(\nu_r) \textrm{ fully lies in the image plane)}, 
\end{align}
which satisfies the second condition in~\eqref{eq:sos}; see proof in supplementary material. 


\subsubsection{GPU acceleration}

The main costs in Algorithm~\ref{alg:cap} are computing the contrast~\eqref{contrast}, contrast upper bound~\eqref{eq:daqiform}, and the line-pixel intersections~\eqref{eq:imgup}. These routines are readily amenable to GPU acceleration via CUDA (this was suggested in~\cite{Liu2020-eo}, but no details or results were reported).


A key step is CUDA memory allocation to reduce the overheads of reallocating memory and copying data between CPU-GPU across the BnB iterations:
\begin{verbatim}
cudaMallocManaged(&event_image, …)
cudaMallocManaged(&event_lines, …) 
cudaMalloc(&events, …)
cudaMemcpy(&events, cpu_events, …)
\end{verbatim}
Summing all pixels (including with squaring the pixels), can be parallelised through the CUDA Thrust library:
\begin{verbatim}
sum_of_pixels = thrust::reduce(event_image_plane, …)
sum_of_square_of_pixels = thrust::transform_reduce(
    event_image_plane, square<double>…,plus<double>…, …)
\end{verbatim}
Finally, the computation of all lines and line-pixel intersections can be parallelised through CUDA device functions:
\begin{verbatim}
get_upper_bound_lines <<< number_of_blocks, block_size >>> 
	    (events, event_lines, …)
get_upper_bound_image <<< number_of_blocks, block_size >>> 
	    (event_lines, event_image_plane, …)
\end{verbatim}
Our CUDA acceleration (on an Nvidia GeForce RTX 2060) resulted in a runtime reduction of $80\%$ for the routines above, relative to pure CPU implementation.

\subsection{Divergence estimation for event stream}

The algorithm discussed in Sec.~\ref{sec:algo} can be used to estimate divergence for a single event batch $\cE$. To estimate the time-varying divergence for the event stream $\cS$, we incrementally take event batches from $\cS$, and estimate the divergence of each batch as discrete samples of the continuous divergence value. In Sec.~\ref{sec:results}, we will report the batch sizes (which affect the sampling frequency) used in our work.

\section{Ventral landing event dataset}\label{sec:datacollection}

Due to the lack of event datasets for ventral landing, we collected our own dataset to benchmark the methods. Our dataset consists of
\begin{itemize}[leftmargin=1em,itemsep=0pt,parsep=0pt,topsep=2pt]
    \item 1 simulated event sequence (\texttt{SurfSim}) of duration 47 seconds.
    \item 7 real event sequences observing 2D prints of landing surfaces (\texttt{Surf2D-X}, where \texttt{X} = \texttt{1} to \texttt{7}) of duration 15 seconds each.
    \item 1 real event sequence observing the 3D print of a landing surface (\texttt{Surf3D}) of duration 15 seconds.
\end{itemize}

\texttt{SurfSim} was produced using v2e~\cite{Hu2021-kl} to process a video that was graphically rendered using PANGU~\cite{pangu_web}, of a camera undergoing pure ventral landing on a lunar-like surface (see Fig.~\ref{fig:vertical_descent_examples}).

\begin{figure}[t]\centering
     \begin{subfigure}[b]{0.37\textwidth}
         \centering
         \includegraphics[width=\textwidth]{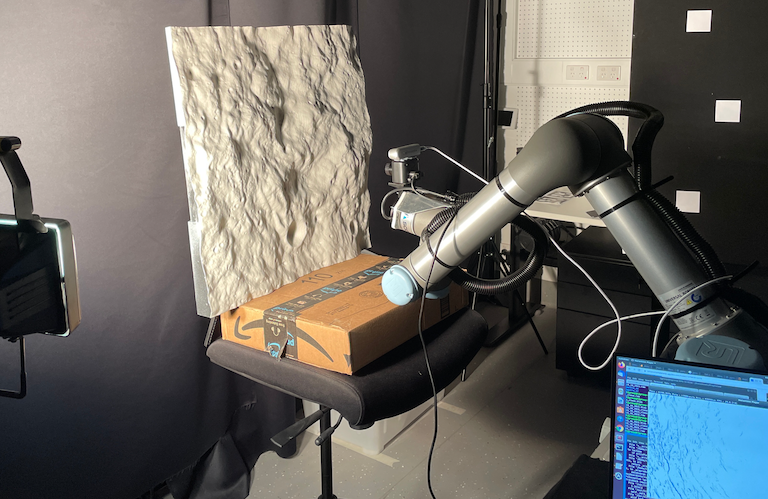}
         \caption{}
         \label{fig:data_collection}
     \end{subfigure}
     \begin{subfigure}[b]{0.24\textwidth}
         \centering
         \includegraphics[width=\textwidth]{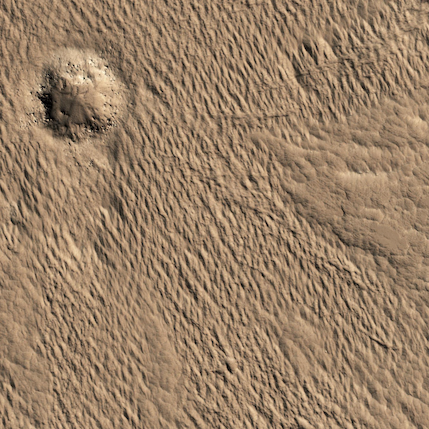}
         \caption{}
         \label{fig:moon_1}
     \end{subfigure}
     \begin{subfigure}[b]{0.24\textwidth}
         \centering
         \includegraphics[width=\textwidth]{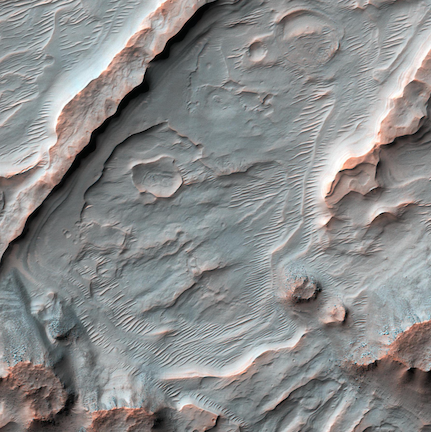}
         \caption{}
         \label{fig:moon_2}
     \end{subfigure}
        \caption{(a) Our setup for recording real event data under (approximate) ventral landing, shown here with the 3D printed lunar surface. Note the only approximately fronto-parallel  surface. (b)(c)~\cite{b_surface_img,c_surface_img} Two of the planetary surface images that were 2D printed. See supplementary material for more details.}
        \label{fig:setup}
\end{figure}

Fig.~\ref{fig:setup} depicts our setup for recording the real event sequences. A UR5 robot arm was employed to execute a controlled linear trajectory towards horizontally facing 2D printed planetary surfaces (Mars, Mercury, \etc) and a 3D printed lunar surface (based on the lunar DEM from NASA/JAXA~\cite{dem_lunar_web}). A Prophesee Gen 4 event camera and Intel RealSense depth sensor were attached to the end effector, while a work lamp provided lighting. Ground truth divergences at $33$ Hz were computed from velocities and depths retrieved from the robot arm and depth camera. Note that, despite using a robot arm, the real sequences do not fully obey the ventral landing assumptions, due to the surfaces not being strictly fronto-parallel to the camera, and the 3D surface not being planar.

\section{Results}\label{sec:results}

We tested our method, \textbf{e}xact \textbf{c}ontrast \textbf{m}aximisation for event-based \textbf{d}ivergence estimation (ECMD), on the dataset described in Sec.~\ref{sec:datacollection}. We compared against: 
\begin{itemize}[leftmargin=1em,itemsep=0pt,parsep=0pt,topsep=2pt]
    \item Local plane fitting (PF) method~\cite{Pijnacker_Hordijk2018-xj};
    \item 2D centroid tracking (CT) method~\cite{Sikorski2021-tf};
    \item Learning-based dense OF method (E-RAFT)~\cite{Gehrig2021-ej}; and
    \item Tracking and grid search-based OF method (SOFAS)~\cite{Stoffregen2018-nm,Stoffregen2019-jc}.
\end{itemize}
Only PF and CT were aimed at divergence estimation, whereas SOFAS and E-RAFT are general OF methods. While there are other event-based OF methods, \eg,~\cite{Almatrafi2020-ot,Gallego2018-tu,Gehrig2019-vc,Liu2018-bs,Pan2020-ju,Ye2018-my}, our list includes the main categories of heuristic, optimisation and learning methods which had publicly available implementations.

\subsubsection{Preprocessing}

All event sequences were subject to radial undistortion and hot pixel removal. Two versions of the data were also produced:
\begin{itemize}[leftmargin=1em,itemsep=0pt,parsep=0pt,topsep=2pt]
    \item \emph{Full resolution}: The original data collected for \texttt{Surf2D-X} and \texttt{Surf3D} by our Prophesee Gen 4 event camera ($1280$x$760$ pixels) and simulated data for \texttt{SurfSim} ($512$x$512$ pixels), with only the processing above.
    \item \emph{Resized and resampled}: The effective image resolution of the camera was resized (by scaling the image coordinates of the events) to $160$x$90$ pixels for \texttt{Surf2D-X} and \texttt{Surf3D}, and $64$x$64$ pixels for \texttt{SurfSim}. The size of each event batch was also reduced by $75\%$ by random sampling.
\end{itemize}



\subsubsection{Implementation and tuning}

For ECMD, the convergence threshold $\gamma$ (see Algorithm~\ref{alg:cap}) was set to 0.025, and an event batch duration of $0.5$ s was used. For PF, we used the implementation available at~\cite{mav_web}. There were 16 hyperparameters that influence the estimated divergence in nontrivial ways, thus we used the default values. Note that PF is an asynchronous algorithm, thus its temporal sampling frequency is not directly comparable to the other methods. We used our own implementation of CT. The two main parameters were the temporal memory and batch duration, which we tuned to to $0.3$ s and $0.1$ s for best performance. For SOFAS, we used the implementation available at~\cite{sofas_web} with batch duration of $0.5$ s. For E-RAFT, we used the implementation available at~\cite{eraft_web}. As there was no network training code available, we used the provided DSEC pre-trained network. E-RAFT was configured to output OF estimates every $0.5$ s.

All methods were executed on a standard machine with an Core i7 CPU and RTX 2060 GPU. See supplementary material for more implementation details.


\subsubsection{Converting OF to divergence}

For E-RAFT and SOFAS, the divergence for each event batch was calculated from the OF estimates by finding the rate of perceived expansion of the OF vectors, as defined in~\cite{Sikorski2021-tf} as
\begin{equation}\label{Sikorski_equation}
    D = \dfrac{1}{P\tau}\sum_{k=1}^P \left[ 1-\dfrac{\lVert FOE+\bp_k+\tau \bv_k \rVert}{\lVert FOE + \bp_k| \rVert}\right],
\end{equation}
where $P$ is the number of OF vectors, $\bv_k$ is the $k$-th OF vector at corresponding pixel location $\bp_k$, and $\tau$ is the duration of the event batch. 





\subsubsection{Comparison metrics}

For each method, we recorded:
\begin{itemize}[leftmargin=1em,itemsep=0pt,parsep=0pt,topsep=2pt]
    \item \emph{Divergence error}: The absolute error (in $\%$) between the estimated divergence and closest-in-time ground truth divergence for each event batch.
    \item \emph{Runtime}: The runtime of the method on the event batch.
\end{itemize}
Since the methods were operated at different temporal resolutions for best performance, to allow comparisons we report results at $0.5$ s time resolution, by taking the average divergence estimate of methods that operated at finer time scales (specifically PF at $0.01$ s and CT at $0.3$ s) over $0.5$ s windows.

\subsection{Qualitative results}

\begin{figure}[tp]
     \centering
     \begin{subfigure}[b]{0.24\textwidth}
         \centering
         \includegraphics[width=\textwidth]{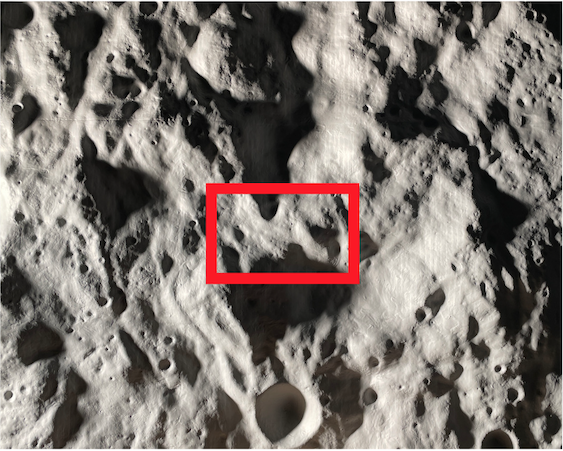}
         \caption{\texttt{Surf2D-1}}
         \label{fig:surf1}
     \end{subfigure}
     \begin{subfigure}[b]{0.24\textwidth}
         \centering
         \includegraphics[width=\textwidth]{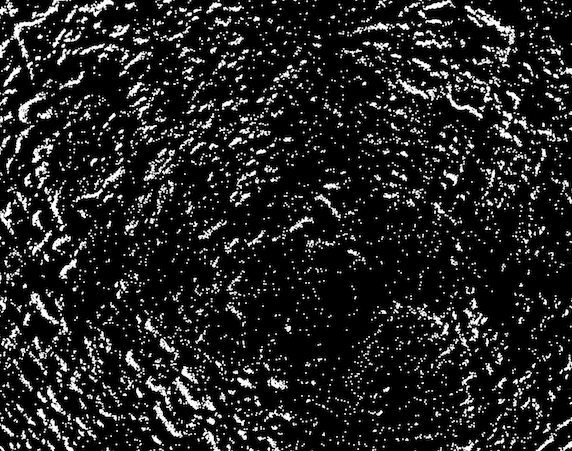}
         \caption{GT (0.033)}
         \label{fig:gt_F}
     \end{subfigure}
     \begin{subfigure}[b]{0.24\textwidth}
         \centering
         \includegraphics[width=\textwidth]{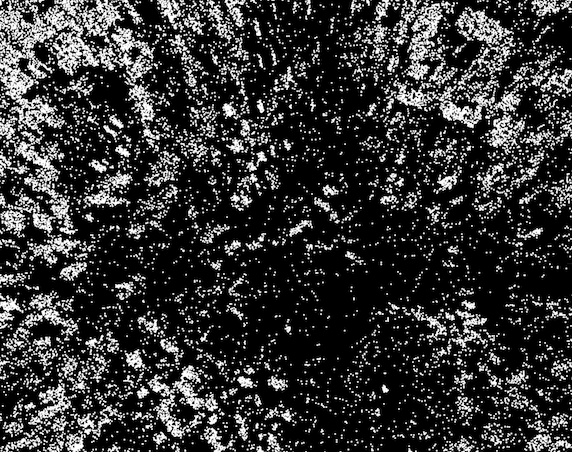}
         \caption{Zero (0.026)}
         \label{fig:zero_f}
     \end{subfigure}
     \begin{subfigure}[b]{0.24\textwidth}
         \centering
         \includegraphics[width=\textwidth]{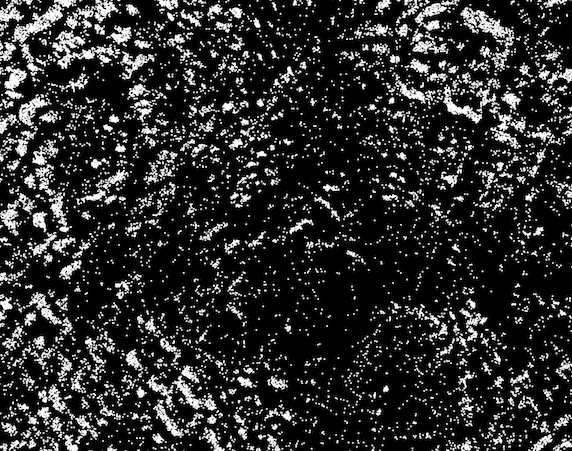}
         \caption{PF (0.028)}
         \label{fig:pf_f}
     \end{subfigure}
     \begin{subfigure}[b]{0.24\textwidth}
         \centering
         \includegraphics[width=\textwidth]{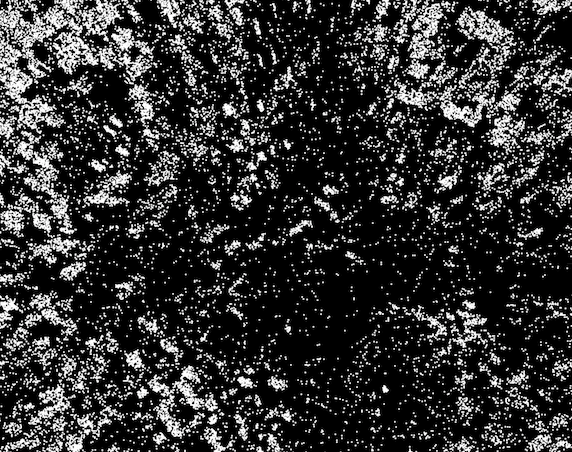}
         \caption{CT (0.026)}
         \label{fig:ct_f}
     \end{subfigure}
     \begin{subfigure}[b]{0.24\textwidth}
         \centering
         \includegraphics[width=\textwidth]{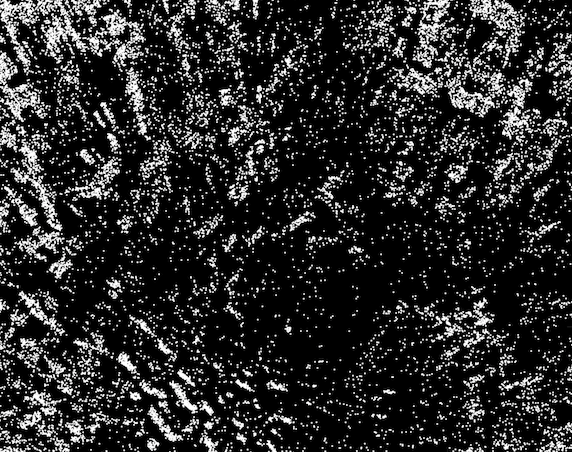}
         \caption{E-RAFT (0.024)}
         \label{fig:eraft_f}
     \end{subfigure}
     \begin{subfigure}[b]{0.24\textwidth}
         \centering
         \includegraphics[width=\textwidth]{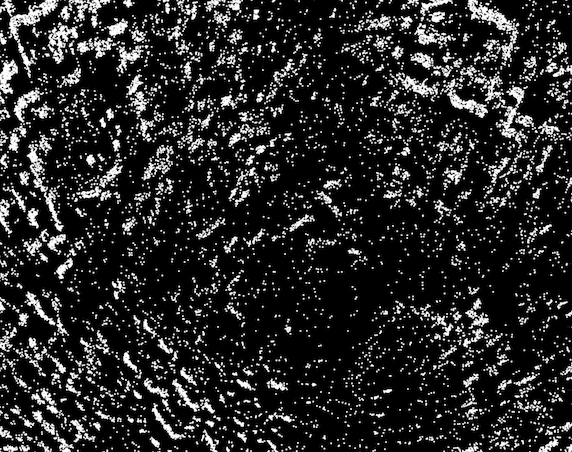}
         \caption{SOFAS (0.027)}
         \label{fig:sofas_f}
     \end{subfigure}
     \begin{subfigure}[b]{0.24\textwidth}
         \centering
         \includegraphics[width=\textwidth]{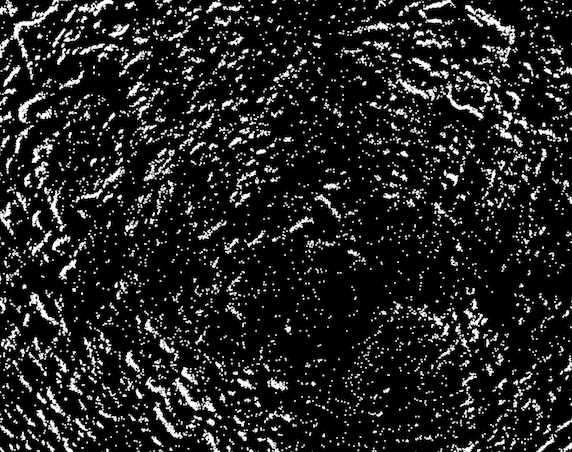}
         \caption{ECMD (0.033)}
         \label{fig:ecmd_f}
     \end{subfigure}

     \begin{subfigure}[b]{0.24\textwidth}
         \centering
         \includegraphics[width=\textwidth]{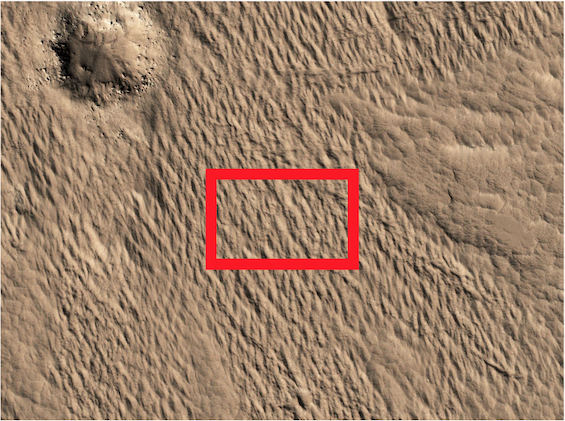}
         \caption{\texttt{Surf2D-6}}
         \label{fig:surf6}
     \end{subfigure}
     \begin{subfigure}[b]{0.24\textwidth}
         \centering
         \includegraphics[width=\textwidth]{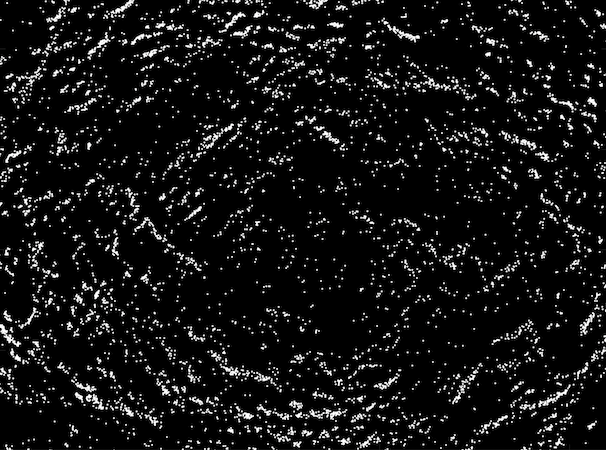}
         \caption{GT (1.82)}
         \label{fig:gt_F}
     \end{subfigure}
     \begin{subfigure}[b]{0.24\textwidth}
         \centering
         \includegraphics[width=\textwidth]{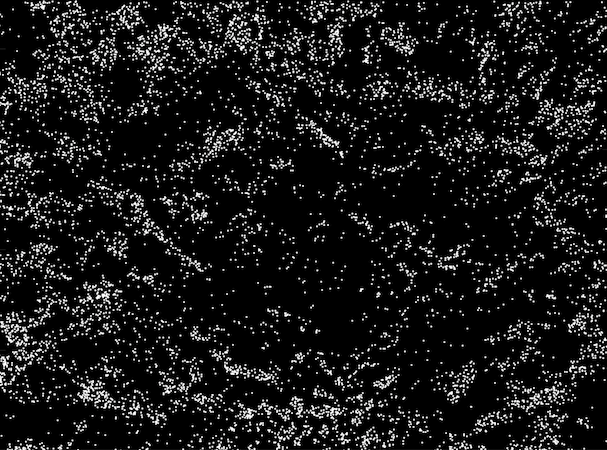}
         \caption{Zero (1.05)}
         \label{fig:zero_f}
     \end{subfigure}
     \begin{subfigure}[b]{0.24\textwidth}
         \centering
         \includegraphics[width=\textwidth]{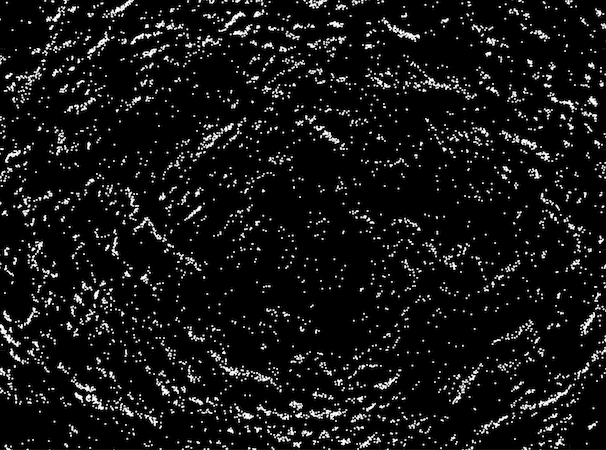}
         \caption{PF (1.8)}
         \label{fig:pf_f}
     \end{subfigure}
     \begin{subfigure}[b]{0.24\textwidth}
         \centering
         \includegraphics[width=\textwidth]{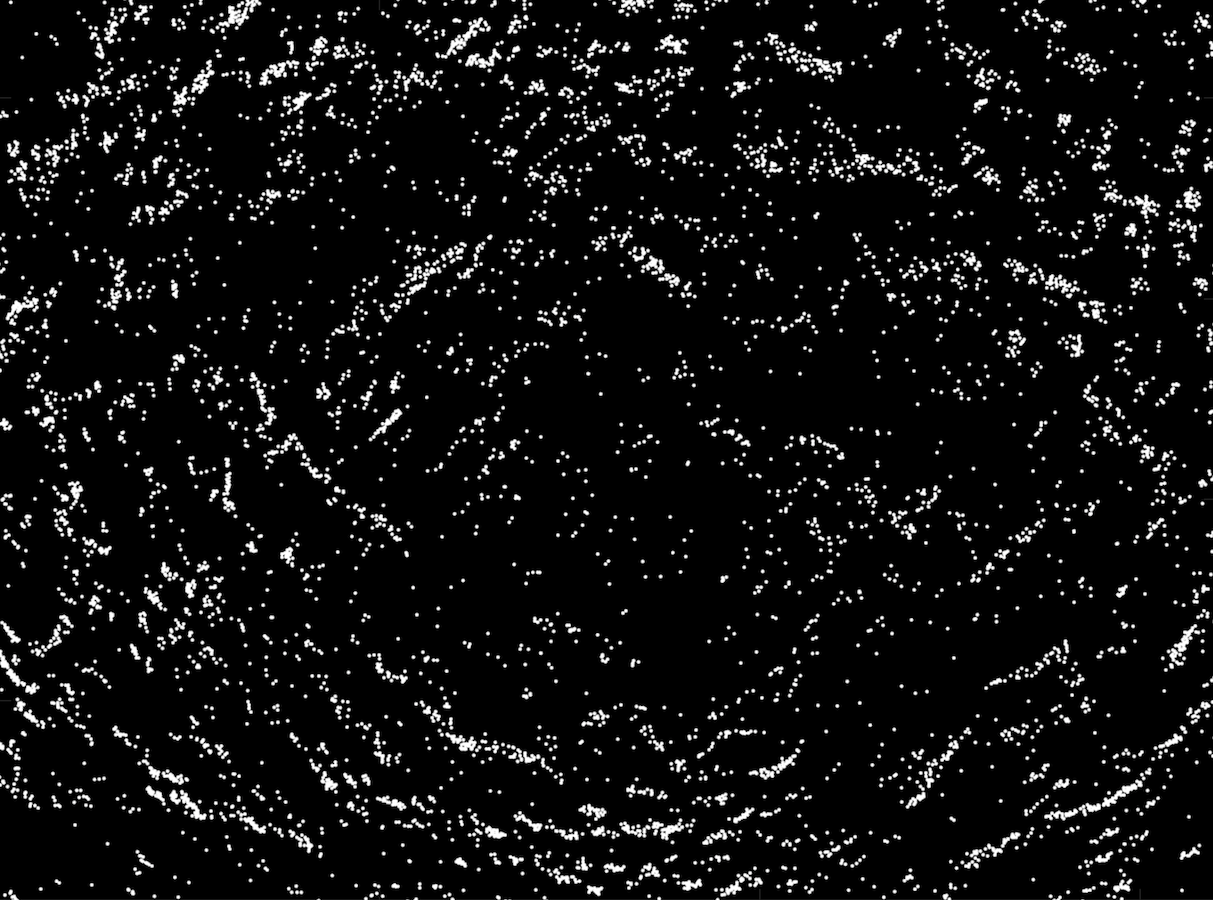}
         \caption{CT (1.64)}
         \label{fig:ct_f}
     \end{subfigure}
     \begin{subfigure}[b]{0.24\textwidth}
         \centering
         \includegraphics[width=\textwidth]{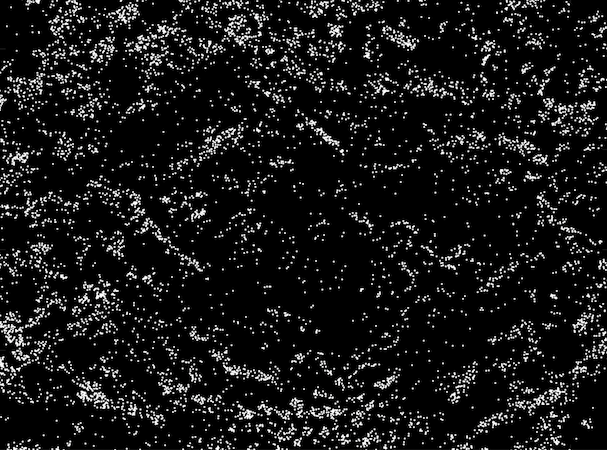}
         \caption{E-RAFT (1.21)}
         \label{fig:eraft_f}
     \end{subfigure}
     \begin{subfigure}[b]{0.24\textwidth}
         \centering
         \includegraphics[width=\textwidth]{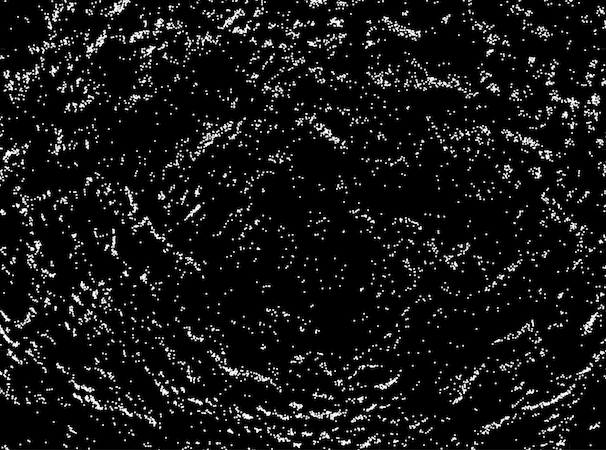}
         \caption{SOFAS (1.64)}
         \label{fig:sofas_f}
     \end{subfigure}
     \begin{subfigure}[b]{0.24\textwidth}
         \centering
         \includegraphics[width=\textwidth]{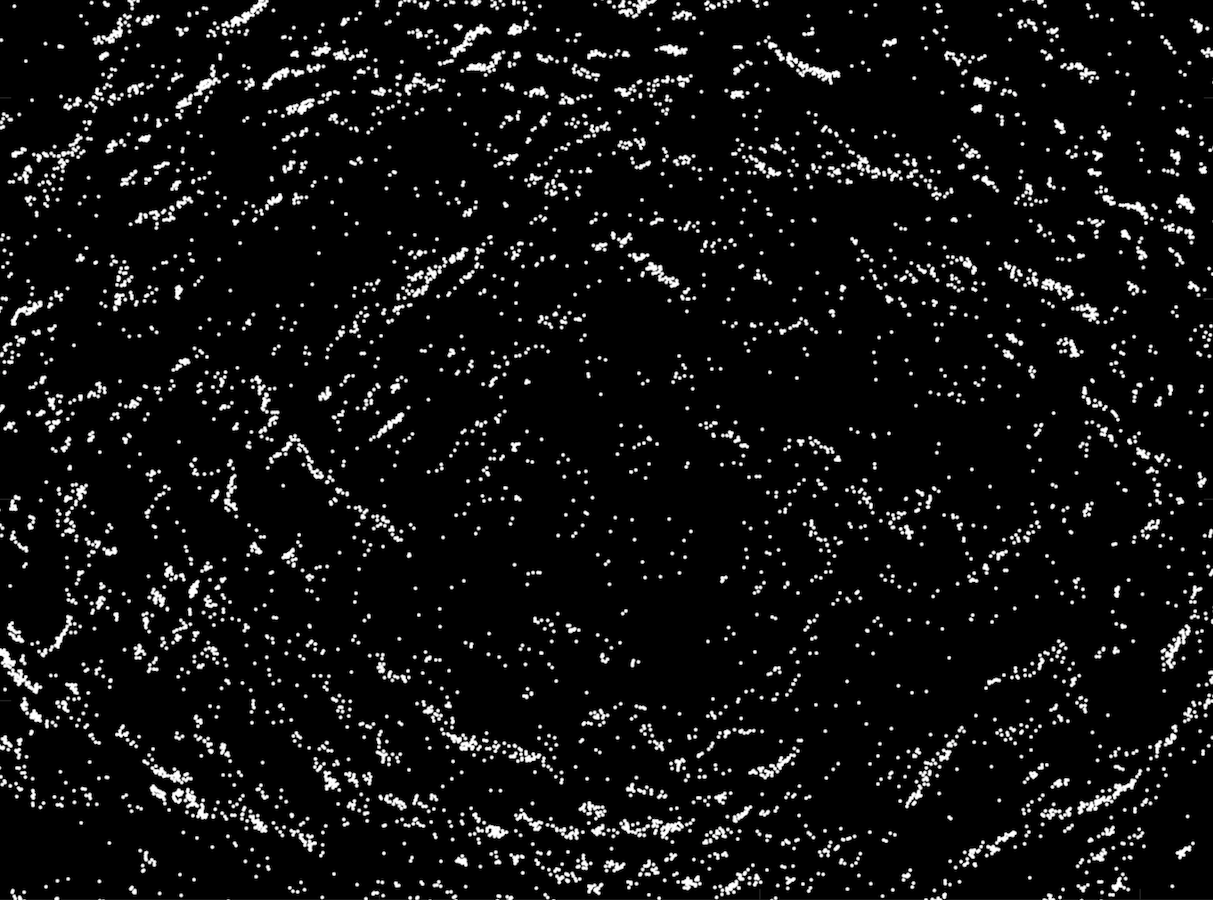}
         \caption{ECMD (1.82)}
         \label{fig:ecmd_f}
     \end{subfigure}
        \caption{(b)-(h) Flow-compensated event images produced for an event batch (full resolution) from \texttt{Surf2D-1}. (j)-(p) Flow-compensated event images produced for an event batch (resized and resampled) from \texttt{Surf2D-6}~\cite{b_surface_img}. GT = Ground truth. Zero = No warp. Numbers in parentheses are the contrast values~\eqref{contrast}.}
        \label{fig:qualitative}
\end{figure}

Fig.~\ref{fig:qualitative} shows motion compensated event images from the estimated $\nu$ across all methods in Sec.~\ref{sec:results}, including event images produced under ground truth (GT) $\nu$, and when $\nu = 0$ (Zero). In these examples, ECMD returned the highest image contrast (close to the contrast of GT).

\subsection{Quantitative results on full resolution sequences}

\begin{figure}[tp]\centering
\includegraphics[width=0.99\textwidth]{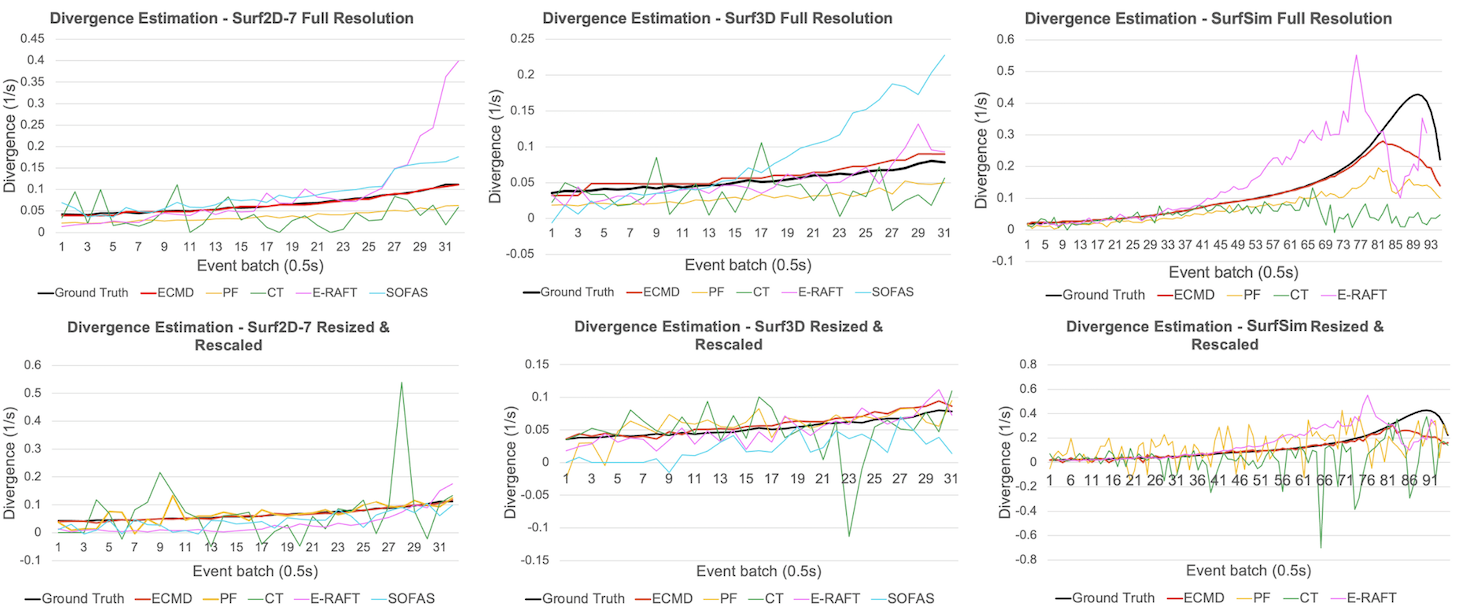}
\caption{Divergence estimates from the methods described in Sec.~\ref{sec:results} for datasets \texttt{Surf2D-7}, \texttt{Surf3D}, and \texttt{SurfSim}. (Top row) Full resolution sequences. (Bottom row) Resized and resampled sequences. Note that SOFAS was not plotted for \texttt{SurfSim} since the divergence estimates were too innacurate (see Table~\ref{table:results_1}).}
\label{fig:divergence_graphs}
\end{figure}

Fig.~\ref{fig:divergence_graphs} (top row) plots the estimated divergence for all methods across 3 of the full resolution event sequences; see supplementary material for the rest of the sequences. Observe that ECMD generally tracked the ground truth divergence well and was the most stable method. The estimates of the other methods were either inaccurate or quite erratic. A common failure point was towards the end of \texttt{SurfSim}; this was when the camera was very close to the surface and the limited resolution of the rendered image did not generate informative events.

Table~\ref{table:results_1} (top) provides average divergence errors and runtimes. Values of the former reflect the trends in Fig.~\ref{fig:divergence_graphs}; across all sequences, ECMD committed the least error relative to the ground truth. The results suggest that, although ECMD was based upon a pure vertical landing assumption, the method was not greatly affected by violations of the assumption in the real event sequences. 

While PF and E-RAFT were more efficient (up to one order of magnitude faster than ECMD), their estimates were rather inaccurate. Despite GPU acceleration, the runtime of ECMD was longer than event batch duration. We will present results on the resized and resampled sequences next.

\begin{table}[tp]
\begin{center}
\arrayrulecolor{black} 
\caption{Divergence estimation results (error and runtime) on event sequences. The final column contains the average error and runtime across all sequences.}
\scriptsize
\label{table:results_1}
\begin{tabular}{ |p{1.2cm}|p{1.2cm}||p{0.8cm}|p{0.8cm}|p{0.8cm}|p{0.8cm}|p{0.8cm}|p{0.8cm}|p{0.8cm}|p{0.8cm}|p{0.8cm}|p{0.8cm}|}
  \hline
  \multicolumn{12}{|c|}{Full resolution event sequences} \\
  \hline 
  \multicolumn{2}{|l|}{Dataset} & Surf-Sim & Surf 2D-1 & Surf 2D-2 & Surf 2D-3 & Surf 2D-4 & Surf 2D-5 & Surf 2D-6 & Surf 2D-7 & Surf-3D & Avg. \\
  \hline
  \multirow{2}{1.2cm}{Absolute average error per event batch(\%)}      
       & ECMD & 8.34 & 13.48 & 7.41 & 6.11 & 12.19 & 10.41 & 5.12 & 3.72 & 12.90 & 8.85\\
       & PF& 37.78 &   39.11 & 44.79 & 44.42 & 41.19 & 42.19 & 43.42 & 44.03 & 25.23 & 40.24\\
       & CT  & 44.33 & 43.86 & 50.27 & 42.96 & 50.07 & 72.80 & 53.73 & 60.30 & 45.85 & 51.57\\
       & E-RAFT & 46.27 & 50.79 &  44.81 & 44.75 &41.12 & 29.90 & 53.53 & 49.19 & 21.82 & 43.46\\
       & SOFAS & 819.33 & 24.85 & 24.65 & 23.63 & 30.67 & 34.80 & 37.05 & 30.88 & 71.54 & 121.93\\
  \hline
  \multirow{2}{1cm}{Average runtime per event batch(s)}      
       & ECMD  & 7.99 & 6.84 & 6.95 & 6.50 & 5.85 & 7.62 & 6.02 &3.01 &2.36 & 5.90\\
       & PF&  0.21 & 0.22 & 0.23 & 0.30 & 0.18 & 0.28 & 0.22 & 0.14 & 0.13 & 0.21\\
       & CT & 35.12 & 31.58 & 35.46 & 41.67 & 28.48 & 48.49 & 42.17 & 21.43 & 20.98 & 33.93\\
       & E-RAFT & 0.24 & 0.25 & 0.25 & 0.24 & 0.25 & 0.25 & 0.24 & 0.25 & 0.24 & 0.25\\
       & SOFAS & 57.80 & 61.46 & 66.43 & 65.79 & 55.62 & 66.23 & 59.32 & 13.27 & 35.83 & 53.53\\
  \hline 
\end{tabular}


\begin{tabular}{ |p{1.2cm}|p{1.2cm}||p{0.8cm}|p{0.8cm}|p{0.8cm}|p{0.8cm}|p{0.8cm}|p{0.8cm}|p{0.8cm}|p{0.8cm}|p{0.8cm}|p{0.8cm}|}
  \hline
  \multicolumn{12}{|c|}{Resized and resampled event sequences} \\
  \hline 
  \multicolumn{2}{|l|}{Dataset} & Surf-Sim & Surf 2D-1 & Surf 2D-2 & Surf 2D-3 & Surf 2D-4 & Surf 2D-5 & Surf 2D-6 & Surf 2D-7 & Surf-3D & Avg. \\
  \hline
  \multirow{2}{1.2cm}{Absolute average error per event batch(\%)}      
       & ECMD  &18.65 & 18.86 & 10.33 & 7.36 & 14.45 & 12.28 & 7.41 & 4.25 & 11.68 &11.70 \\
       & PF& 146.76&   58.66 & 32.79 & 50.49 & 52.66 & 43.80 & 57.36 & 30.61 & 31.64 & 56.09\\
       & CT & 102.20&143.99 & 51.83 & 73.35 & 40.57 & 81.72 & 89.83 & 96.80 & 43.47 & 80.42 \\
       & E-RAFT & 47.40& 86.40  & 72.31  & 88.62 & 81.67 &71.87 &91.26 &66.79 &21.19 & 69.72\\
       & SOFAS & 1688.21 & 60.10 & 50.49 & 56.29 & 67.73 & 47.87 & 24.42 & 42.98 & 63.51 &233.51  \\
  \hline
  \multirow{2}{1cm}{Average runtime per event batch(s)}      
       & ECMD  & 0.20 &0.31 & 0.29 & 0.32 & 0.34 & 0.34 & 0.30 & 0.28 & 0.23 & 0.29\\
       & PF& 0.12&  0.10 & 0.10 & 0.10 & 0.17 & 0.07 & 0.12 & 0.10 & 0.06 & 0.10\\
       & CT  & 7.81 &20.65 & 22.09 & 30.59 & 32.79 & 13.99 & 27.38 & 11.63 & 15.33 & 20.25\\
       & E-RAFT & 0.24 & 0.24  & 0.25 & 0.24 & 0.24 &0.25 &0.25 & 0.25 &0.24 &  0.24\\
       & SOFAS & 6.14 & 36.18 & 39.06 & 39.54 & 38.67 & 34.63 & 40.13 & 36.96 & 25.27  &32.95 \\
  \hline 
\end{tabular}
\end{center}
\end{table}

\subsection{Quantitative results on resized and resampled sequences}

Fig.~\ref{fig:divergence_graphs} (bottom row) plots the estimated divergence for all methods across 3 of the resized and resampled event sequences; see supplementary material for the rest of the sequences. Table~\ref{table:results_1} (bottom) provides average  errors and runtimes.

The results show that the accuracy of all methods generally deteriorated due to the lower image resolution and event subsampling. However, the impact on the accuracy of ECMD has been relatively small, with a difference in average error of within $3\%$ from the full resolution results. Though PF was still the fastest method, the average runtime of ECMD has reduced by about $95\%$. The average runtime of ECMD was $0.29$ s, which was below the batch duration.

\section{Conclusions and future work}

This paper proposes exact contrast maximisation for event-based divergence estimation (ECMD) for guiding ventral descent. Compared to other divergence and OF methods, ECMD demonstrated stable and accurate divergence estimates across our ventral landing dataset. Violations to the ECMD assumption on a pure ventral landing did not greatly affect the divergence estimates. Through GPU acceleration and downsampling, ECMD achieved competitive runtimes.

Future work includes integrating ECMD into a closed loop system~\cite{Pijnacker_Hordijk2018-xj,Sikorski2021-tf}, and comparing GPU-enabled ECMD to SNN event-based OF methods.

\section{Acknowledgements}

Sofia Mcleod was supported by the Australian Government RTP Scholarship. Tat-Jun Chin is SmartSat CRC Professorial Chair of Sentient Satellites.






\clearpage
%
%
\bibliographystyle{splncs04}
\bibliography{ms}

\end{document}


\pagestyle{headings}
\mainmatter
\def\ECCVSubNumber{175}  

\title{Supplementary Material:\\Globally Optimal Event-Based Divergence Estimation for Ventral Landing} 

\titlerunning{Globally Optimal Event-Based Divergence Estimation for Ventral Landing}
%

\author{{Sofia McLeod}\inst{1}\index{McLeod, Sofia} \and
{Gabriele Meoni}\inst{2} \and
{Dario Izzo}\inst{2}\and
{Anne Mergy}\inst{2} \and
{Daqi Liu}\inst{1}\and
{Yasir Latif}\inst{1}\and
{Ian Reid}\inst{1}\and
{Tat-Jun Chin}\inst{1}
}

%
\authorrunning{S. McLeod et al.}
%
\institute{School of Computer Science, The University of Adelaide \email{@adelaide.edu.au} \and
Advanced Concepts Team, European Space Research and Technology Centre,
Keplerlaan 1, 2201 AZ Noordwijk, The Netherlands
\email{@esa.int}}
\maketitle

\section{Divergence plots for the rest of the dataset}

\begin{figure}[t]
     \centering
     \begin{subfigure}[b]{0.45\textwidth}
         \centering
         \includegraphics[width=\textwidth]{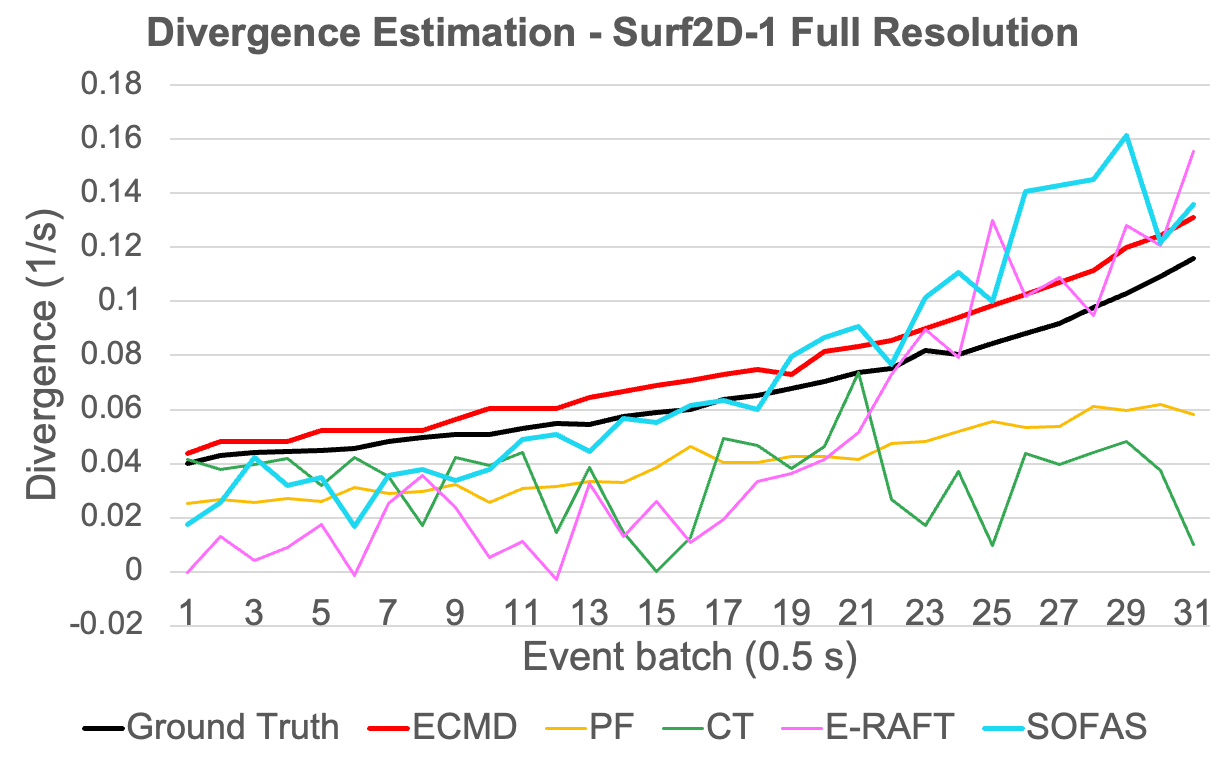}
         \label{fig:full1}
     \end{subfigure}
     \begin{subfigure}[b]{0.45\textwidth}
         \centering
         \includegraphics[width=\textwidth]{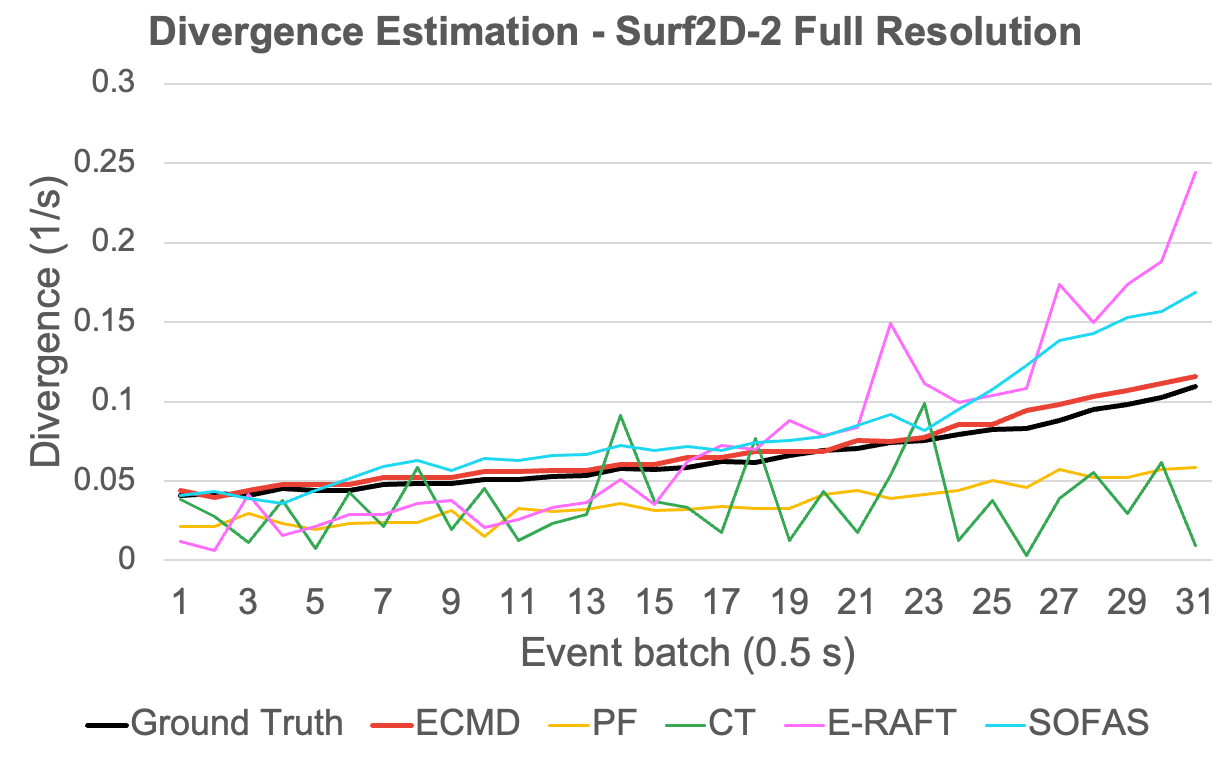}
         \label{fig:full2}
     \end{subfigure}
     \begin{subfigure}[b]{0.45\textwidth}
         \centering
         \includegraphics[width=\textwidth]{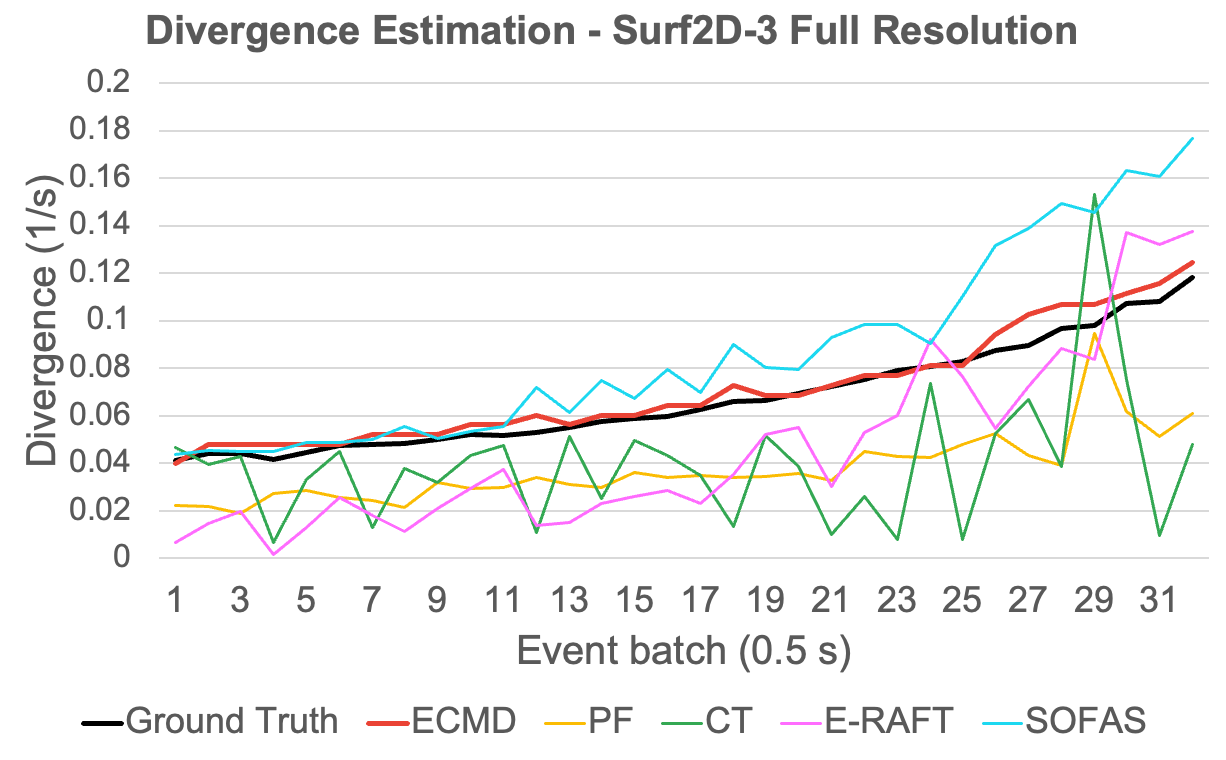}
         \label{fig:full3}
     \end{subfigure}
     \begin{subfigure}[b]{0.45\textwidth}
         \centering
         \includegraphics[width=\textwidth]{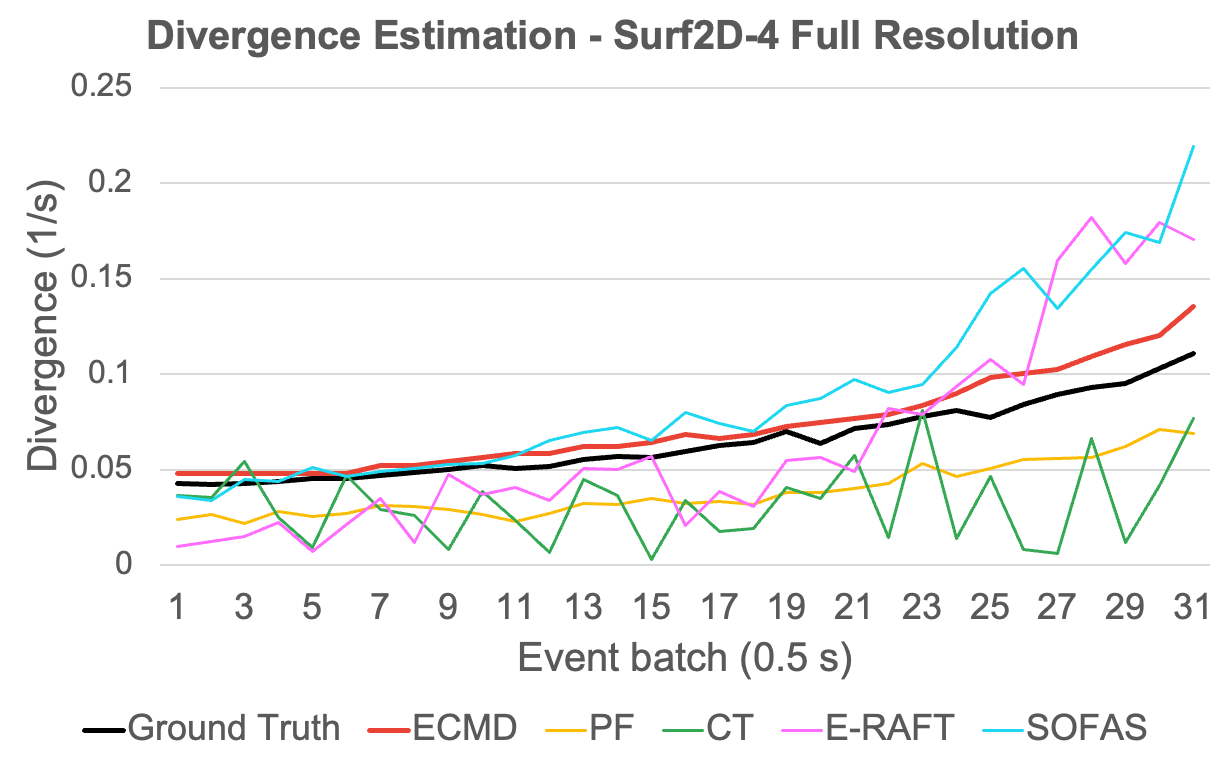}
         \label{fig:full4}
     \end{subfigure}
     \begin{subfigure}[b]{0.45\textwidth}
         \centering
         \includegraphics[width=\textwidth]{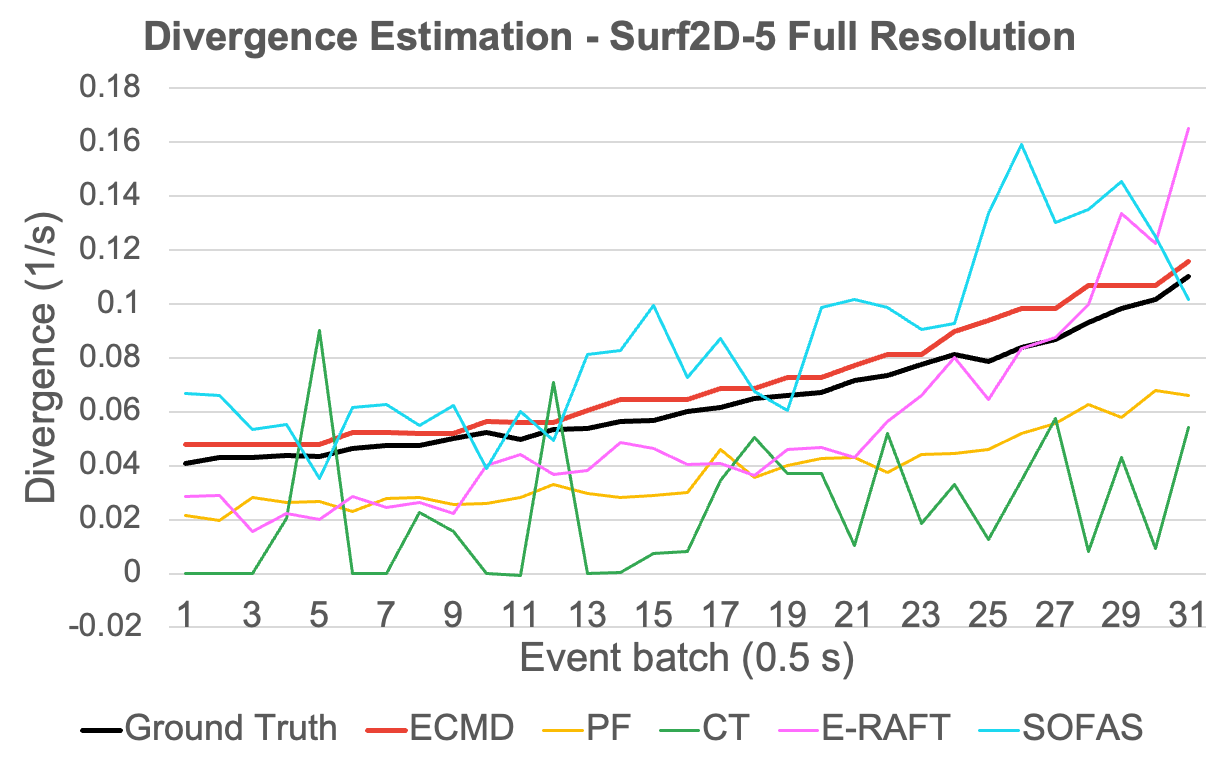}
         \label{fig:full5}
     \end{subfigure}
     \begin{subfigure}[b]{0.45\textwidth}
         \centering
         \includegraphics[width=\textwidth]{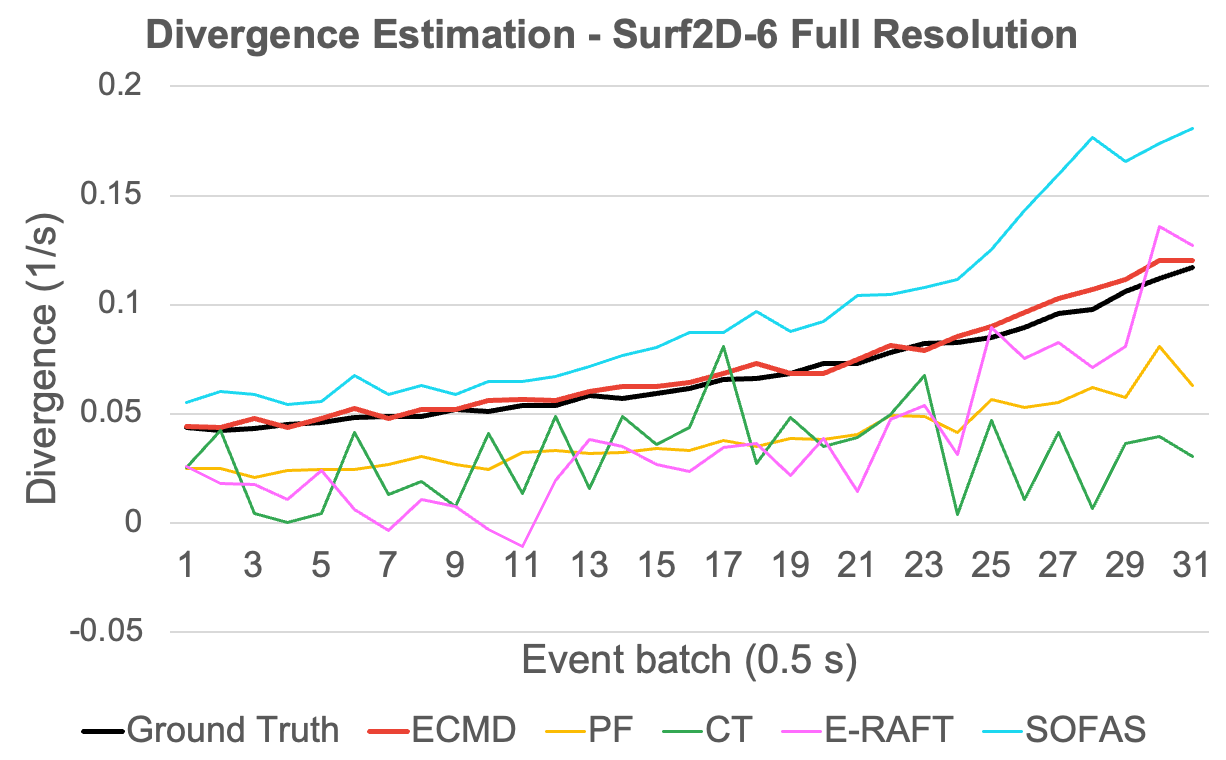}
         \label{fig:full6}
     \end{subfigure}
     
     \vspace{-1em}
        \caption{Divergence estimates from the methods described in Sec.~6 of the main paper the remaining full resolution sequences.}
        \label{fig:full_res}
\end{figure}

\begin{figure}[t]
     \centering
     \begin{subfigure}[b]{0.45\textwidth}
         \centering
         \includegraphics[width=\textwidth]{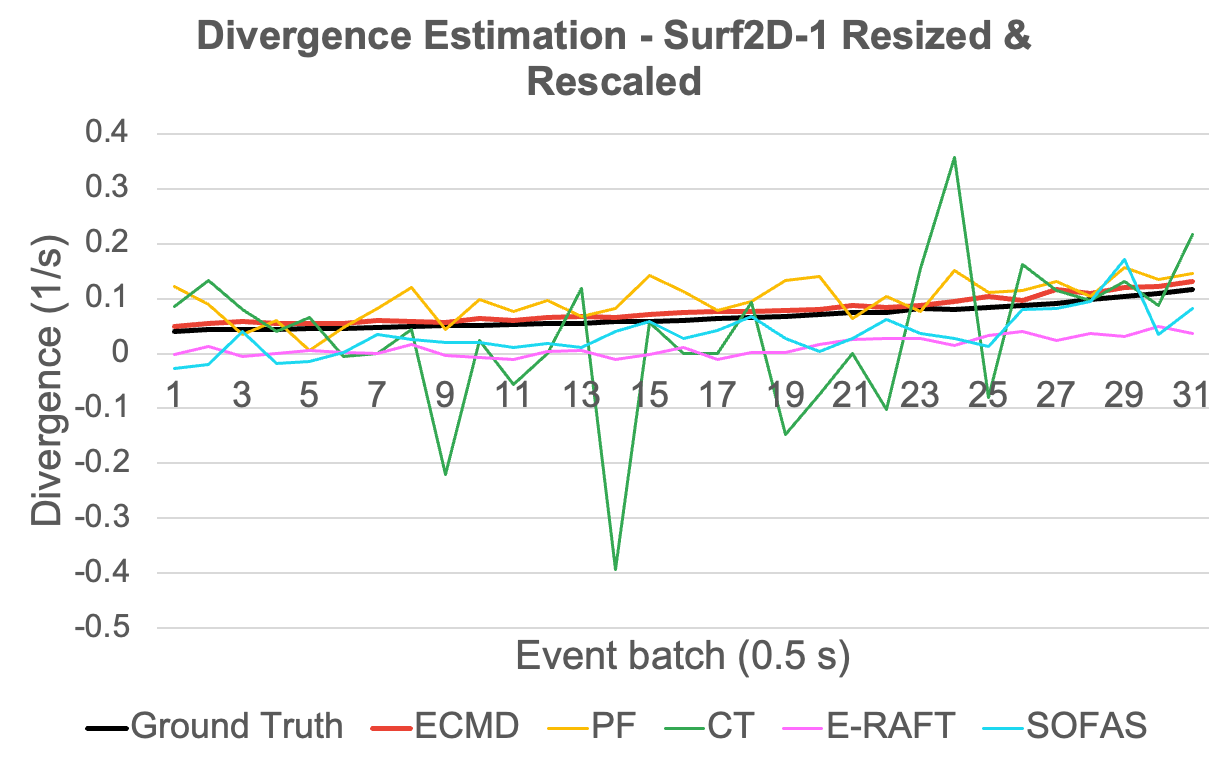}
         \label{fig:resize1}
     \end{subfigure}
     \begin{subfigure}[b]{0.45\textwidth}
         \centering
         \includegraphics[width=\textwidth]{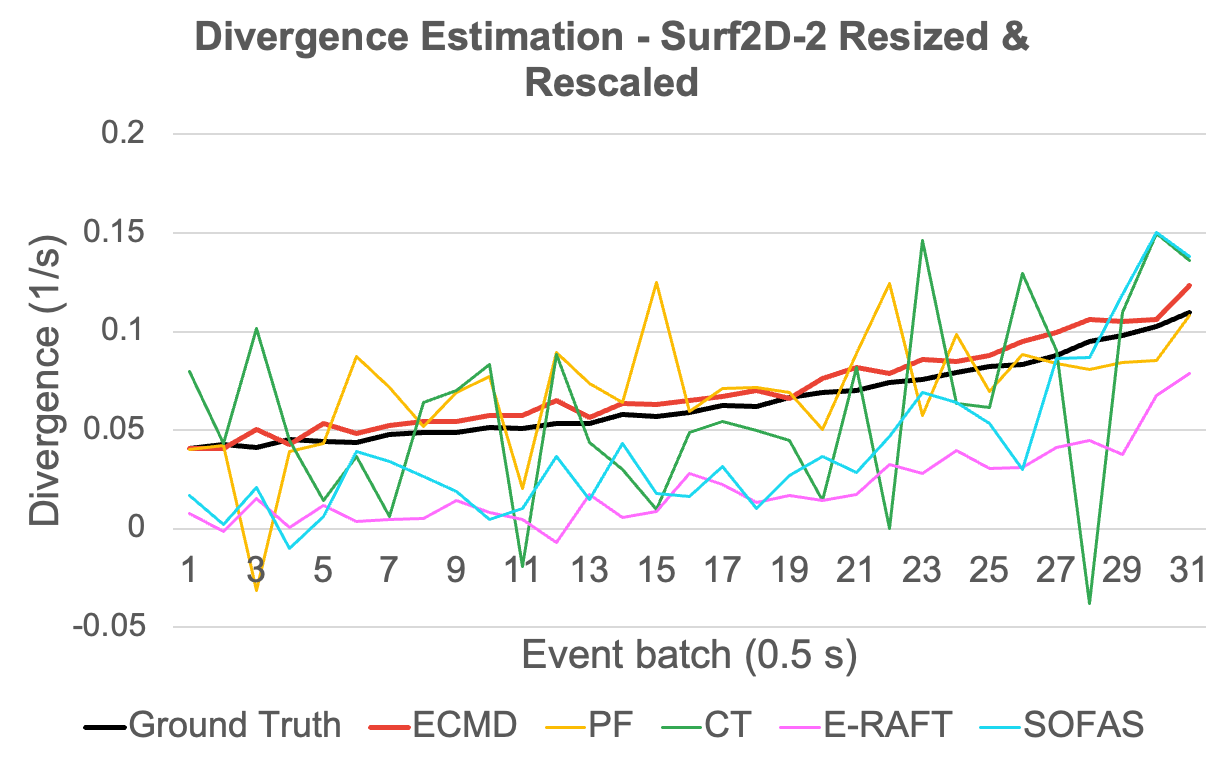}
         \label{fig:resize2}
     \end{subfigure}
     \begin{subfigure}[b]{0.45\textwidth}
         \centering
         \includegraphics[width=\textwidth]{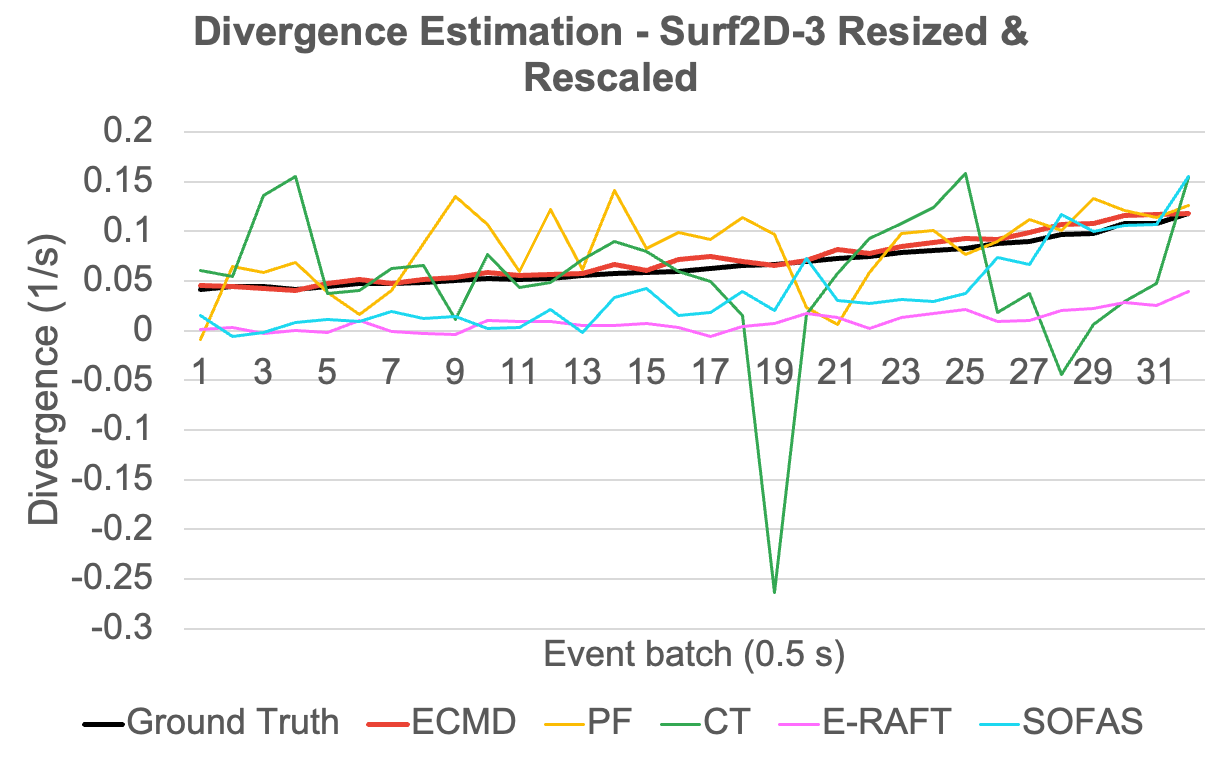}
         \label{fig:resize3}
     \end{subfigure}
     \begin{subfigure}[b]{0.45\textwidth}
         \centering
         \includegraphics[width=\textwidth]{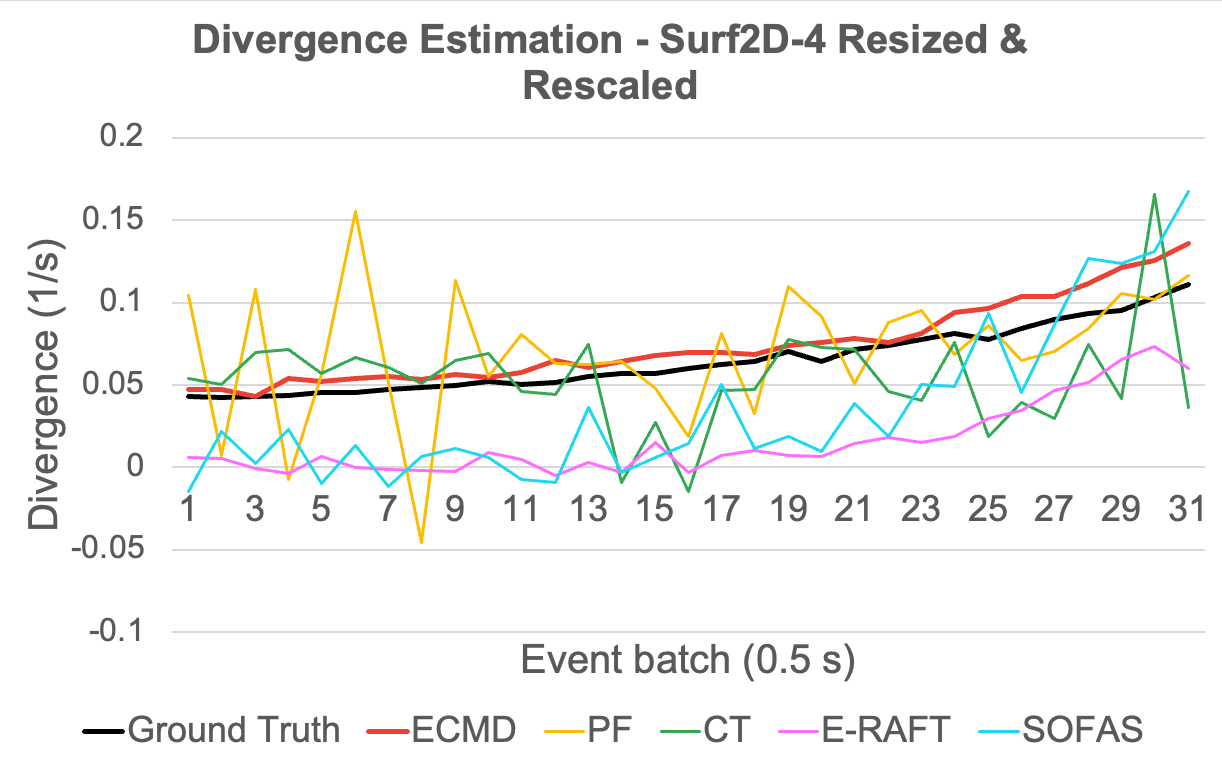}
         \label{fig:resize4}
     \end{subfigure}
     \begin{subfigure}[b]{0.45\textwidth}
         \centering
         \includegraphics[width=\textwidth]{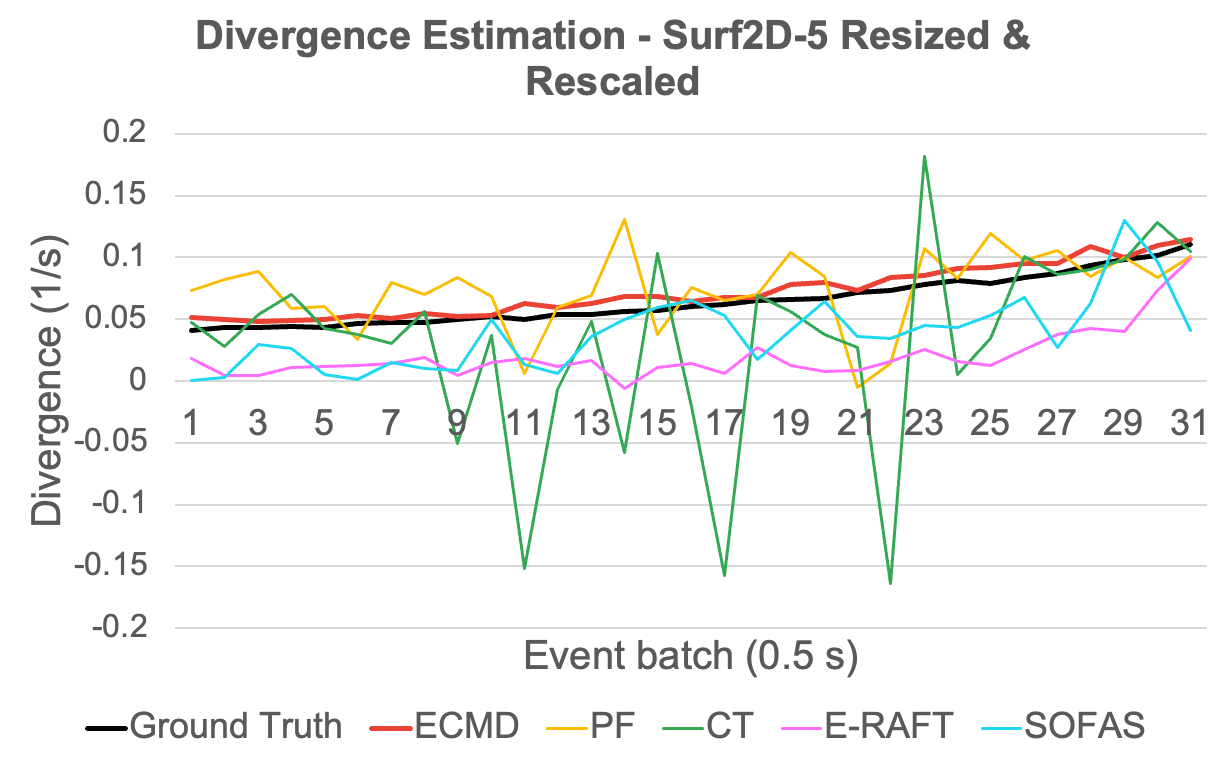}
         \label{fig:resize5}
     \end{subfigure}
     \begin{subfigure}[b]{0.45\textwidth}
         \centering
         \includegraphics[width=\textwidth]{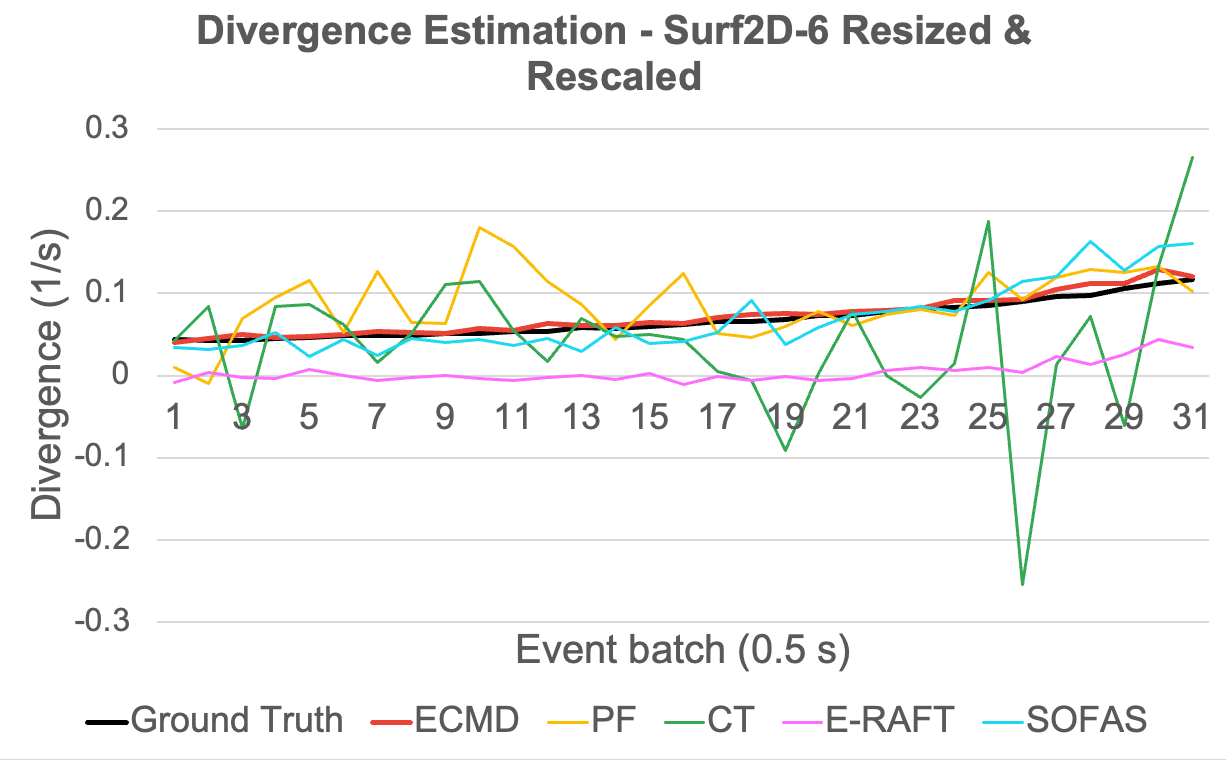}
         \label{fig:resize6}
     \end{subfigure}
     
     \vspace{-1em}
        \caption{Divergence estimates from the methods described in Sec. 6 of the main paper for the remaining resized and resampled sequences.}
        \label{fig:resize}
\end{figure}

Figs.~\ref{fig:full_res} and~\ref{fig:resize} plot the estimated divergences from all methods described in Sec. 6 of the main paper, across the remaining event sequences in the full resolution, and resized and resampled datasets respectively.




\section{More details on data collection}

\begin{figure}[t]
     \centering
     \begin{subfigure}[b]{0.57\textwidth}
         \centering
         \includegraphics[width=\textwidth]{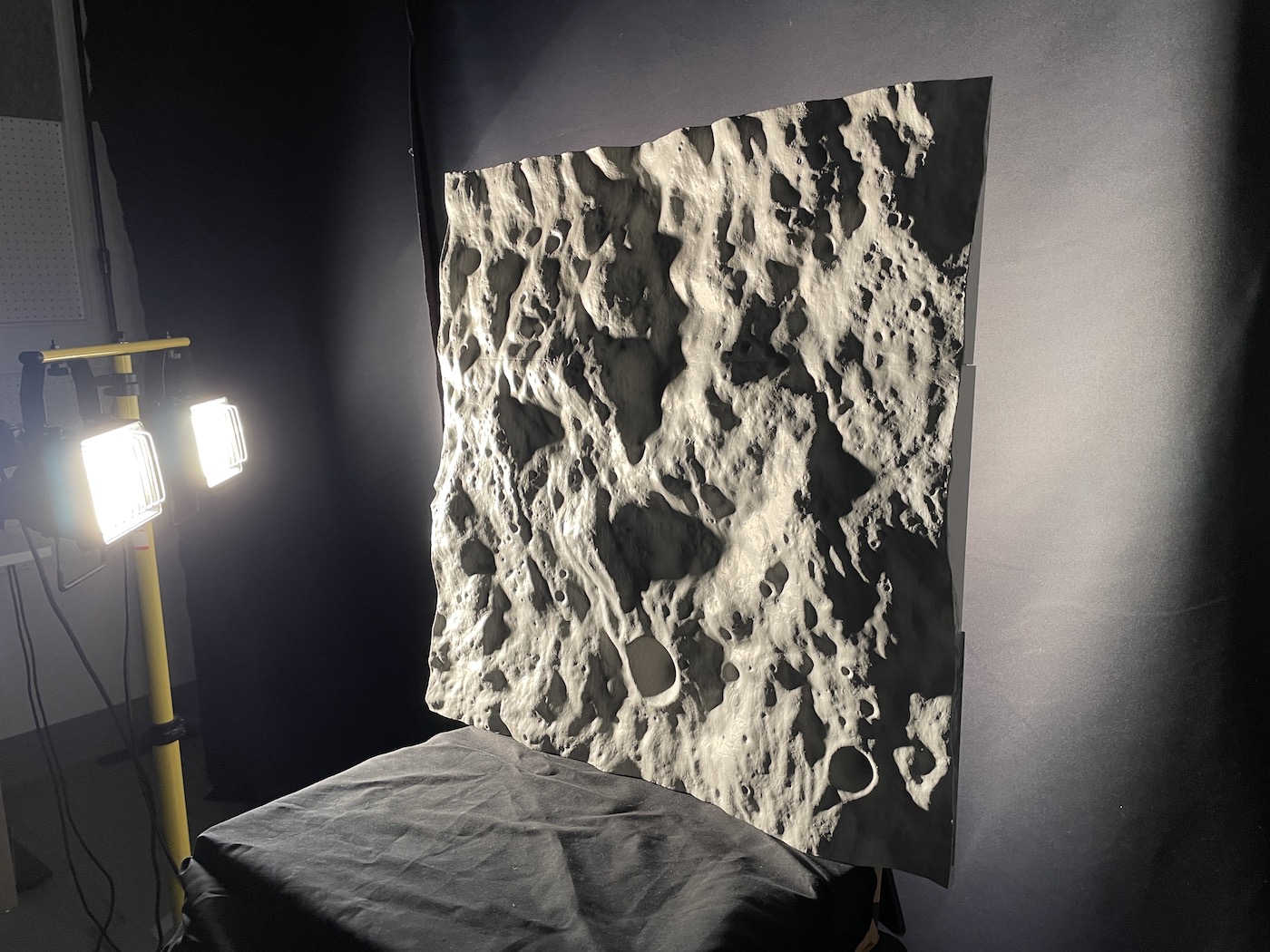}
         \caption{}
         \label{fig:setup}
     \end{subfigure}
     \begin{subfigure}[b]{0.35\textwidth}
         \centering
         \includegraphics[width=\textwidth]{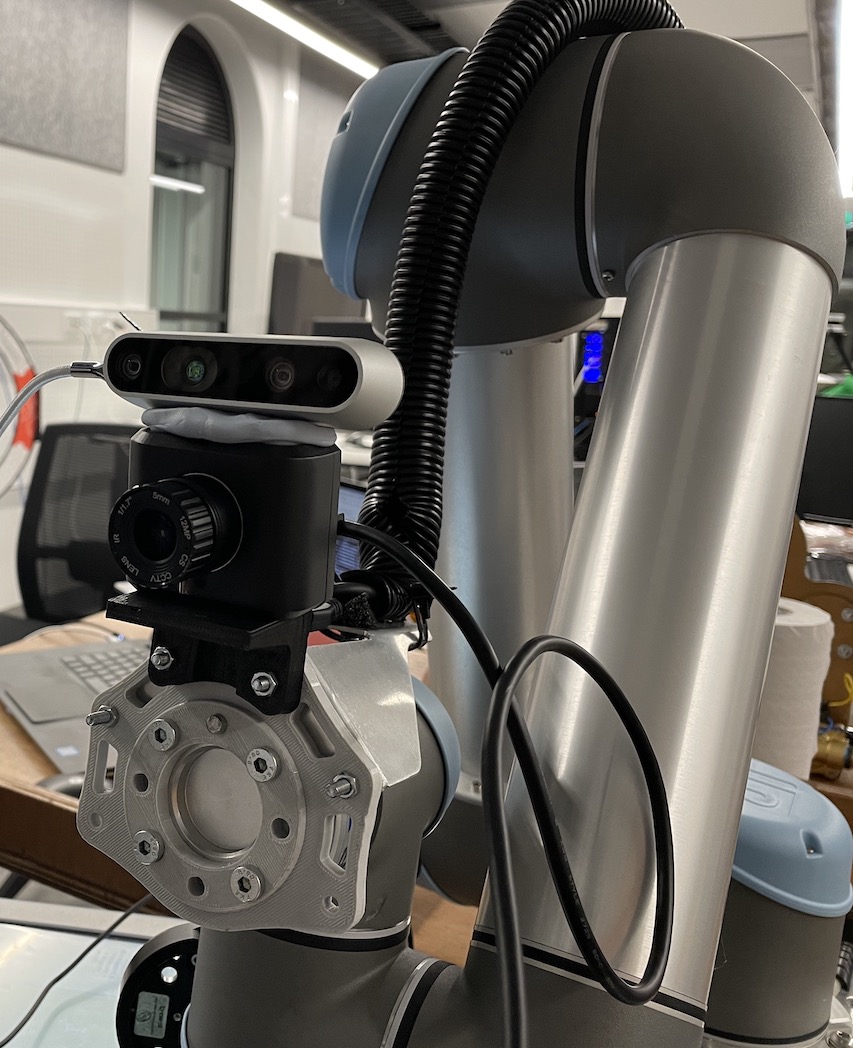}
         \caption{}
         \label{fig:camera_setup}
     \end{subfigure}
     \begin{subfigure}[b]{0.7\textwidth}
         \centering
         \includegraphics[width=\textwidth]{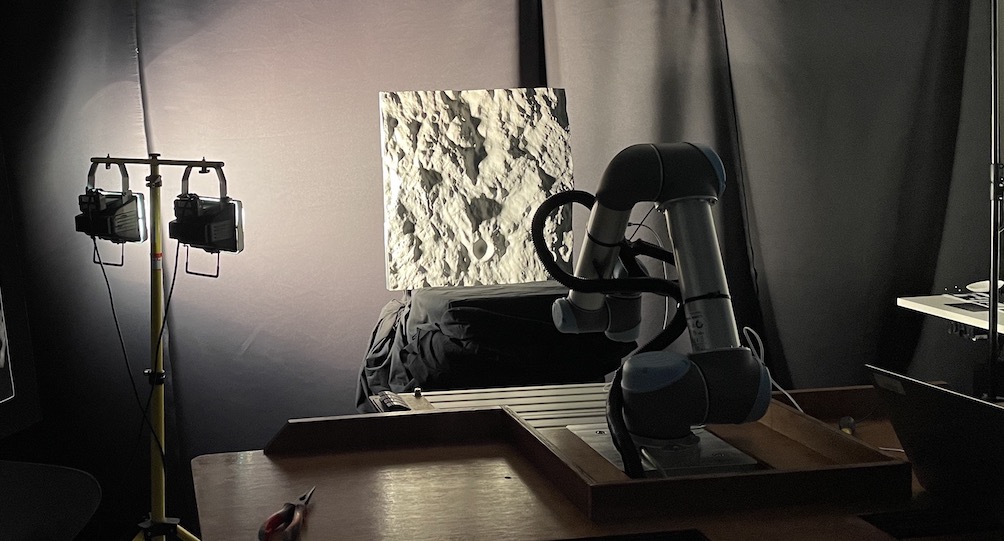}
         \caption{}
         \label{fig:lighting}
     \end{subfigure}
     \vspace{-1em}
        \caption{(a) \texttt{Surf3D} illuminated by work lamp. (b) Prophesee Gen 4 event camera and Intel RealSense depth sensor mounted to the end effector of the UR5 robot arm. (c) Data collection setup: UR5 robot arm (mounted to a table) executing a controlled linear trajectory towards \texttt{Surf3D}.}
        \label{fig:data_collection}
\end{figure}

The ventral descent event data sequences were gathered using the setup depicted in Fig.~\ref{fig:data_collection}.  The UR5 robot was programmed to travel with a linear trajectory towards a 2D or 3D (Fig.~\ref{fig:setup}) lunar or planetary surface. A Prophesee Gen 4 event camera and Intel RealSense depth sensor were attached to the end effector of the UR5 arm (Fig.~\ref{fig:camera_setup}).  The event camera recorded events and the RealSense camera recorded distances to the surface, throughout the travelled trajectory.  Simultaneously, the UR5's ground truth pose was recorded through internal sensors, which was used to determine the linear velocity of the trajectory. Both velocity and distance estimates were used to retrieve the ground truth divergence of the event camera throughout the ventral ``descent'' towards the surface.


\section{Completing the proof of upper bound}

This section provides more details for Sec.~4.3 on the main paper. Specifically, we defined the contrast of an event image in the main paper
\begin{align}\label{eq:imgcont}
    C(\nu) = \dfrac{1}{M}\sum_{\bu} (H(\bu;\nu) - \mu (\nu))^2,
\end{align}
whereby following~\cite[Sec. 3.3]{Liu2020-eo}, \eqref{eq:imgcont} can be rewritten as
\begin{align}\label{eq:daqiform4}
    C(\nu) = \dfrac{1}{M}\sum_{\bu} H(\bu;\nu)^2 - \mu (\nu)^2.
\end{align}
Given the constraint on the upper bound
\begin{align}\label{eq:daqiform1}
    \bar{C}(\cV) \geq \max_{\nu \in \cV} C(\nu),
\end{align}
the upper bound was defined in the main paper as
\begin{align}
    \bar{C}(\cV) = \frac{1}{M}\bar{S}(\cV) - \underline{\mu}(\cV)^2.
\end{align}
We defined in the main paper the upper bound image
\begin{align}\label{eq:imgup}
    \bar{H}(\bu; \cV) = \sum_{i=1}^N \mathbb{I}(f_i(\nu_\ell) \leftrightarrow f_i(\nu_r) \textrm{ intersects with pixel } \bu),
\end{align}
where $\cV = [\nu_\ell,\nu_r]$, and $f_i(\nu_\ell) \leftrightarrow f_i(\nu_r)$ is the line on the image plane joining the points $f_i(\nu_\ell)$ and $f_i(\nu_r)$. We aim to establish
\begin{align}\label{eq:imguppbound}
    \bar{H}(\bu; \cV) \ge H(\bu; \nu), \;\;\;\; \forall \nu \in \cV,
\end{align}
with equality achieved when $\cV$ reaches a single point $\nu$ in the limit. By definition of the event-based radial warp
\begin{equation}\label{eq:warping_function}
    f_i(\nu) = \begin{bmatrix} x_i \\ y_i \end{bmatrix} + \begin{bmatrix} \dfrac{x_i(Z_0+\nu t_i)}{Z_0+\nu \tau} - x_i, & \dfrac{y_i(Z_0+\nu t_i)}{Z_0+\nu \tau} - y_i \end{bmatrix}^T,
\end{equation}
for each $i$ we must have that
\begin{align}
    f_i(\nu)~~\textrm{lies on}~~f_i(\nu_{\ell}) \leftrightarrow f_i(\nu_{r})
\end{align}
for all $\nu \in [\nu_\ell,\nu_r]$. Therefore,
\begin{align}
    \mathbb{I} (f_i(\nu) \textrm{ lies in pixel } \bu) \le \mathbb{I}(f_i(\nu_\ell) \leftrightarrow f_i(\nu_r) \textrm{ intersects with pixel } \bu)
\end{align}
thus establishing~\eqref{eq:imguppbound}. When $\cV$ converges to a single point $\nu$, the line $f_i(\nu_\ell) \leftrightarrow f_i(\nu_r)$ also converges to the point $f_i(\nu)$ thus leading to equality in~\eqref{eq:imguppbound}.



As described in Sec.~4.3 in the main paper, we defined 
\begin{align}\label{eq:mulower}
    \underline{\mu}(\cV) = \dfrac{1}{M}\sum_{i=1}^N\mathbb{I} (f_i(\nu_\ell) \leftrightarrow f_i(\nu_r) \textrm{ fully lies in the image plane)}. 
\end{align}
We aim to establish
\begin{align}\label{eq:sos2}
    \underline{\mu}(\cV) \le \min_{\nu \in \cV} \mu(\nu),
\end{align}
with equality achieved when $\cV$ is singleton. For each $i$, if $f_i(\nu_\ell) \leftrightarrow f_i(\nu_r)$ fully lies in the image plane, then the $i$-th event contributes no less than $1$ to the summation in~\eqref{eq:sos2} under all $\nu \in \cV$. On the other hand, if $f_i(\nu_\ell) \leftrightarrow f_i(\nu_r)$ does not fully lie in the image plane, there is $\nu \in \cV$ that will cause the $i$-th event to contribute $0$ to the summation. This implies condition~\eqref{eq:sos2}. When $\cV$ converges to a single point $\nu$, the line $f_i(\nu_\ell) \leftrightarrow f_i(\nu_r)$ also converges to the point $f_i(\nu)$ thus leading to equality in~\eqref{eq:sos2}.


\section{More implementation details}

The SOFAS~\cite{Stoffregen2018-nm,Stoffregen2019-jc} implementation available at~\cite{sofas_web} is capable of estimating a single optic flow vector across an entire event stream, and therefore in it's pure form, is not able to directly estimate divergence.  A way to compensate for this would be to recover the OF through event-based motion segmentation~\cite{Stoffregen2019-me}, and then use (21) from the main paper to find the divergence.  However, as this code is not publicly available, we instead estimate the OF across patches of an event batch, where each event patch contains a subset of events bounded by the patch's x and y dimensions.  After all OF vectors are generated for each patch, the divergence can be estimated using (21) from the main paper, where $\bp_k$ is the centre pixel of the patch.  A 10x10 patch grid over the event image plane (i.e. 100 patches) obtains best divergence estimates for our dataset.

\bibliographystyle{splncs04}
\bibliography{supplement}